\documentclass[pdflatex,sn-mathphys]{sn-jnl}

\jyear{2022}%

\theoremstyle{thmstyleone}%
\newtheorem{theorem}{Theorem}
\newtheorem{proposition}[theorem]{Proposition}%

\theoremstyle{thmstyletwo}%
\newtheorem{example}{Example}%
\newtheorem{remark}{Remark}%
\newtheorem{assumption}{Assumption}%
\newtheorem{corollary}{Corollary}%
\theoremstyle{thmstylethree}%
\newtheorem{definition}{Definition}%

\begin{document}

\title[ELDD]{Exploring the Learning Difficulty of Data: Theory and Measure}


\author{\fnm{Zhu} \sur{Weiyao}}\email{weiyaozhu042@outlook.com}

\author{\fnm{Wu} \sur{Ou*}}\email{wuou@tju.edu.cn}

\author{\fnm{Su} \sur{Fengguang}}

\author{\fnm{Deng} \sur{Yingjun}}

\affil{\orgdiv{National Center for Applied Mathematics}, \orgname{Tianjin University}, \orgaddress{\street{92 Weijin Road}, \city{Tianjin}, \postcode{300072}, \country{China}}}


\abstract{As learning difficulty is crucial for machine learning (e.g., difficulty-based weighting learning strategies), previous literature has proposed a number of learning difficulty measures. However, no comprehensive investigation for learning difficulty is available to date, resulting in that nearly all existing measures are heuristically defined without a rigorous theoretical foundation. In addition, there is no formal definition of easy and hard samples even though they are crucial in many studies. This study attempts to conduct a pilot theoretical study for learning difficulty of samples. First, a theoretical definition of learning difficulty is proposed on the basis of the bias-variance trade-off theory on generalization error. Theoretical definitions of easy and hard samples are established on the basis of the proposed definition. A practical measure of learning difficulty is given as well inspired by the formal definition. Second, the properties for learning difficulty-based weighting strategies are explored. Subsequently, several classical weighting methods in machine learning can be well explained on account of explored properties. Third, the proposed measure is evaluated to verify its reasonability and superiority in terms of several main difficulty factors. The comparison in these experiments indicates that the proposed measure significantly outperforms the other measures throughout the experiments.}




\maketitle

\section{Introduction}\label{sec1}
\indent

The partition of training data into different subsets according to their learning difficulties and adoption of separate learning schemes (e.g., weighting) are proven to be useful in many learning tasks~\cite{b2,b3,b31,b16}. The learning difficulty investigated in this study refers to the degrees of easy or hard to learn of training samples in a given learning task. Although learning difficulty has no formal and consensus definition, it has been widely discussed and utilized in previous machine learning literature, including noise-aware, curriculum, and metric learning. 
\indent

Numerous methods are proposed to measure the learning difficulty of a training sample. The most common practice is to leverage the training output (e.g., loss and the predicted value on the true category) of a sample to construct the measurements. In Self-paced Learning (SPL)~\cite{b3,b15}, the training loss is used to determine whether a sample is easy or not, and easy samples are first learned. We assume that $p_{i,y_i}$ is the prediction on the ground-truth category for a training sample $x_i$. In object detection, the value of $(1-p_{i,y_i})$ is used to indicate the learning difficulty for $x_i$~\cite{b16} . Given that the training output in an epoch may be unreliable, some methods utilize the average training output of a sample during the training to measure the difficulty. Huang et al.~\cite{b17} designed a cyclic training procedure, and the model is trained from under-fitting to over-fitting in one cycle. The average training loss in the whole cyclic procedure is used as the noisy indicator for a training sample. Feng et al.~\cite{b18} utilized the magnitude of the loss gradient to measure the learning difficulty of a training sample. A large gradient magnitude indicates a high degree of difficulty.
\indent

 Due to lack of a theoretical basis, different learning difficulty measures are based on different heuristic cues or empirical observations, resulting that each measure usually only suits specific application scenarios. A clearer understanding of the essence of a sample's learning difficulty can at least facilitate explaining difficulty-based weighting methods and designing more effective learning difficulty measures. However, we are still far from concluding that we have a comprehensive understanding of learning difficulty:
\begin{itemize}
\item[(1)] There is no formal definition of the learning difficulty of a sample. Different studies exhibit different understandings of learning difficulty. A one-sided understanding usually results in a biased measure.  
\item[(2)] There is no formal definition of the easy and hard samples in a learning task. In most existing studies, easy and hard samples are heuristically judged. Consequently, it is nearly impossible to conduct a theoretical analysis for difficulty-based weighting strategies with existing heuristic considerations. 
\item[(3)] There has been few experimental studies particularly on the learning difficulty measure. Most studies only refer to the noisy learning or uncertainty settings. An extensive empirical evaluation under different settings is useful for the understanding the learning difficulty.
\end{itemize}
\indent

This study attempts to establish a preliminary theoretical definition for learning difficulty on the basis of the basic machine learning theory, namely bias-variance trade-off theory, for generalization errors of samples. The proposed definition leverages the optimal model complexity of a sample to capture its learning difficulty, inspired by previous observations that easy samples are modeled by simple patterns~\cite{b74}. The definitions of easy, medium and hard samples are subsequently proposed based on our theoretical definition. As it is nearly infeasible to calculate the optimal model complexity, a practical approach is proposed to approximate the theoretical definition. Then, the properties for difficulty-based weighting learning methods are explored to support the rationality of the theoretical definitions. Several typical weighting strategies, including SPL and Focal loss, are theoretically analyzed
and explained. The results of the extensive experiments verify the superiority of the proposed measure.

Our contributions are summarized as follows:

\begin{itemize}
    \item An attempt in theoretical definition of learning difficulty is made based on the basic machine learning theory, namely, bias-variance trade-off. To our knowledge, this work is the first step on the formal description of learning difficulty. 
    \item Formal definitions of easy and hard samples are established. As far as we are aware, this is the first attempt on this formalization.
    \item A measure of learning difficulty is proposed and is attested to be rational and exceptional.
    \item The theoretical properties of difficulty-aware weighting strategies in machine learning are explored, and the theoretical explanations based on definition of learning difficulty are presented for several typical weighting methods, which enhance understanding over effective strategies.
\end{itemize}
\indent

\section{Related Work}\label{sec2}

\subsection{Learning Difficulty Measurement}
\indent

Learning difficulty is considered as an intrinsic property of data in machine learning~\cite{b2,b8}. Existing measurements are usually based on heuristic cues or inspirations, and they can be divided into the following main categories:
\begin{itemize}
    \item Loss-based measurement. This category directly uses the loss as the measure. Most measures fall into this category because it is simple yet effective in various learning tasks. Some methods~\cite{b3} directly utilize the loss in one epoch as the degree of difficulty. Accordingly, the degrees for the same samples vary in different epochs. Some others utilize the average loss~\cite{b45} during the partial or whole training procedure for measurement.
    \item Cross-validation-based measurement. This category adopts a cross-validation strategy~\cite{b19}. For example, five-fold cross-validation is performed, and the whole cross-validation is repeated ten times. Consequently, each training sample receives ten predictions. The value of error predictions is used as the indicator of difficulty.
    \item Uncertainty-based measurement. This category uses the (model) uncertainty of a sample to measure the difficulty. Aguilar et al.~\cite{b7} identified hard samples based on the epistemic uncertainty (also known as the model uncertainty). They leveraged the Bayesian Neural Network~\cite{b20} to infer the epistemic uncertainty.
    \item Margin-based measurement. This category uses the margin (distance) of a sample to the underlying decision surface as the measurement. The rationale is that a small margin denotes a large difficulty~\cite{b3, b4}. 
    \item Gradient-based measurement. This category uses the loss gradient of a sample to measure the difficulty. Agarwal and Hooker~\cite{b6} proposed the variance of gradients (VOG) across different epochs to rank data from difficult to easy. They considered that samples with high VOG values are far more difficult for the model to learn. Santiago et al.~\cite{b46} applied the norm of the gradients to measure the difficulty, and high norms indicate large difficulty for learning.
\end{itemize}
\indent

The above-mentioned categories are highly correlated. For example, margin-based measurement is indeed a loss-based one when margin-based loss (e.g., hinge loss) is used.

\subsection{Noisy-label Learning}
\indent

Noisy labels are inevitable even in benchmark data sets~~\cite{b17,b21,b22,b23}. Various methods are explored to detect noisy labels. Existing noise detection methods are usually based on the information used for learning difficulty measurements, such as loss and gradient, because samples with noisy labels are usually considered as quite hard samples. Some studies model the generation process of the noisy labels to detect them~\cite{b14,b24}. A recent survey can be referred to~\cite{b25}.

\subsection{Curriculum Learning}
\indent 

Curriculum learning~\cite{b47} draws lessons from the human learning process, which begins with the simplest and progresses to more difficult courses. Easier samples should be learned at the beginning of a learning process and gradually advance towards harder samples. SPL pertains to curriculum learning and the difficulty is measured by loss. 


\subsection{Uncertainty-aware Learning}
\indent

Uncertainty in learning mainly refers to aleatoric uncertainty and epistemic uncertainty. Aleatoric uncertainty is also called data uncertainty.
A related work~\cite{b59} wisely estimated uncertainty in labeling, which is also recognized as aleatoric uncertainty, since accurate and consistent labeling has high unsureness in real-world. To reduce the influence of high-uncertainty samples, they lower the weights of samples with high density and label entropy. Epistemic uncertainty is also called model uncertainty. It occurs when there is no fixed annotation for a given training sample in some learning tasks. The predictive entropy~\cite{b1} and Bayesian Neural Network~\cite{b20} have been used to measure epistemic uncertainty.

\section{Theoretical Definition of Learning Difficulty}\label{sec3}
\indent

Existing learning difficulty measurements mentioned in Section 2.1 are empirically utilized in diverse situations. Despite the effect emerges under difficulty-based learning schemes, there is still a lot of room of improvement. Because existing measurements are proposed heuristically without a theoretical basis. Moreover, incentives of difference between learning difficulty of samples are multifarious. Existing measurements are mainly of unilateral considerations, which could not cover the majority incentives. 

Arpit et al.~\cite{b74} gave a descriptive definition for easy (as well as hard) samples that ``easier examples are explained by some simple patterns, which are reliably learned within the first epoch of training". This definition implies that easy samples can be modeled by simple models, which motivates us to build a theoretical description with model complexity. Model complexity is a key concept in the classical bias-variance trade-off theory, which is the basis of machine learning and is about the generalization error of a learning task. 

\subsection{Bias-Variance Trade-Off for Generalization Error}\label{AA}
\indent

Bias-variance trade-off is a basic theory for the qualitative analysis of the generalization error in machine learning~\cite{b37}. It is initially constructed on regression and mean square error (MSE) is used~\cite{b37}. The features and the label of a sample are seen as two random variables, and are denoted as $x$ and $y$ respectively. Realizations of a sample are in form of $\lbrace (x_i , y_i) \rbrace$. We assume that $x$ and $y$ conform to the joint distribution $P(x,y)$, where $(x,y) \in \Omega$ with $\Omega = \Omega_X \times \Omega_Y = \lbrace (x,y) \vert x \in \Omega_X, y \in \Omega_Y \rbrace $ . Let $T$ be a random training set and $\lambda_{h} \in R_{\lambda}$ be the hyper-parameter(s), where $R_{\lambda}$ is the feasible region\footnote{Note that the hyper-parameter should locate in a feasible region. For example, if $\lambda_{h}$ is the learning rate, then $\lambda_{h} < 0$ is meaningless.}. Simplify $(x,y)$ as $x$ in the rest of paper. Given a basic learner $f$ and a ﬁxed value of $\lambda_h$, a model can be trained on $T$ and is denoted by $f(x;T,\lambda_{h})$. The generalization error over different realizations of the training samples~\cite{b16} is
\begin{equation}\label{1}
    Err(\lambda_{h} )= \mathbb{E}_{x\in \Omega} \mathbb{E}_T [ \|y-f(x;T,\lambda_{h} )\|_2^2].
\end{equation}
Eq.~(\ref{1}) can be factorized into
\begin{equation}\label{2}
      Err(\lambda_{h} )  = \mathbb{E}_{x\in \Omega} [ \|y-\overline{f}(x;\lambda_{h} )\|_2^2]+  \mathbb{E}_{x\in \Omega} \mathbb{E}_T [ \|\overline{f}(x;\lambda_{h} )-f(x;T,\lambda_{h} )\|_2^2] + \delta_e \  ,
\end{equation}
where $\bar{f}(x;\lambda_h) = \mathbb{E}_T[f(x;T,\lambda_h)]$, and $\delta_e$ is known as the irreducible noise, and is independent of the basic learner and $\lambda_{h}$. The ﬁrst and the second terms of the right side of Eq.~(\ref{2}) are the learning bias and variance terms, respectively, shown as follows:
\begin{equation}\label{3}
        Bias^2(\lambda_{h} ) = \mathbb{E}_{x\in \Omega} [ \|y-\overline{f}(x;\lambda_{h} )\|_2^2].
\end{equation}
\begin{equation}\label{4}
        Var(\lambda_{h} ) =\mathbb{E}_{x\in \Omega} \mathbb{E}_T [ \|\overline{f}(x;\lambda_{h} )-f(x;T,\lambda_{h} )\|_2^2].
\end{equation}

In classiﬁcation, the above derivation becomes complex~\cite{b5}. Nevertheless, the following expression holds, with $BiasT$ and $VarT$ denoting the bias and the variance terms respectively:
\begin{equation}\label{5}
    Err(\lambda_{h}) = BiasT(\lambda_{h}) + VarT(\lambda_{h}) + \delta_e.
\end{equation}

Variable $y$ is categorical in classification. We suppose that $x$ is continuous in order to consider in the total space, and facilitate the inference for classification. Therefore, the generalization error for a region $\Omega^r \subset \Omega$ s.t. $\Omega^r = \Omega_X^r \times \Omega_Y = \lbrace (x,y)\vert x \in \Omega_X^r, y \in \Omega_Y \rbrace $ is defined as 
        \begin{equation}
            Err(\Omega^r, \lambda_{h}) = \sum_{y \in \Omega_Y}   P(y) \int_{x \in \Omega_X^r} \mathbb{E}_T [ l(y,f(x;T,\lambda_{h} ))] p(x\vert y)\mathrm{d}x,
        \end{equation}
where $l(\cdot ,\cdot)$ measures the error between the label and a prediction. $P(y)$ signifies the probability when the label equals to $y$ and $p(x\vert y)$ is the conditional probability density function of $x$ when the label equals to $y$. Denote $m(\cdot)$ as the model complexity defined by a function of parameters of a speciﬁc model $f(x;T,\lambda_{h})$ trained on a sampled training set $T$ with $\lambda_h$. Let $c$ be the expectation of the model complexities given $\lambda_h$. It is defined as follows:
\begin{equation}\label{7}
    c= \mathbb{E}_T [m(f(x;T,\lambda_{h}))].
\end{equation}
In the rest of the paper, the model complexity expectation is briefly termed as ``model complexity". $c$ depends on the base network (e.g., AlexNet, Transformer, and ResNet-34), $\lambda_{h}$ (e.g., learning rate and the maximum learning epoch), and the distribution of $T$. When $f$ and the distribution are ﬁxed, $c$ is the function of $\lambda_h$, i.e.,
\begin{equation}\label{8}
    c = g(\lambda_{h}),\quad \lambda_{h} \in R_{\lambda}.
\end{equation}
In this study, the base network is assumed to be ﬁxed. Therefore, $Err(\lambda_{h})$ can be seen as the function of $c$ according to Eqs. (\ref{1}) and (\ref{8}).
The following widely accepted assumption$\footnote{Most professional books and papers explicitly or implicitly apply this assumption without giving a strict proof. Some recent studies point out that the variance curve is not increasing any more in some cases~\cite{b16}. However, the structures of base models in these studies are also varied. Meanwhile, the structures of base models in this study are assumed to be ﬁxed.}$  holds for both regression and classiﬁcation.

\begin{assumption}\label{assump1}The bias term is a decreasing function of the expectation of model complexity $c$, whereas the variance term is an increasing function of $c$ when the basic learner is ﬁxed. The generalization error decreases first and then increases.
\end{assumption}

\begin{figure}
\centerline{\includegraphics[scale=0.5]{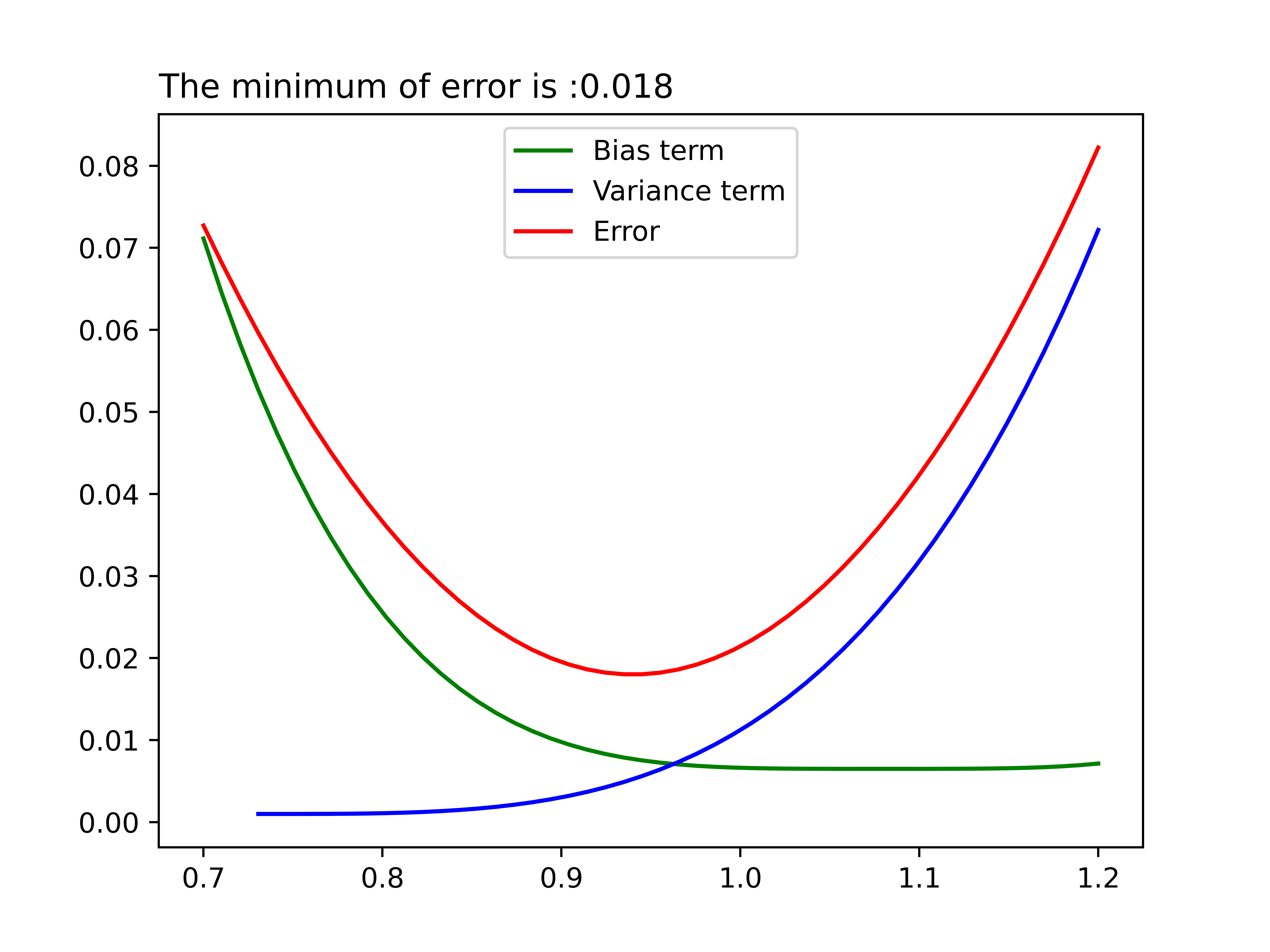}}\vspace{-0.1in}
\caption{Illustration of bias-variance trade-off under non-linear regression.}\vspace{-0.1in}
\label{fig15}
\end{figure}

An example of non-linear regression learning is utilized to empirically support Assumption \ref{assump1}. 
\begin{example}
4000 realizations of random variable $x$ are sampled uniformly from $[0,5]$. The true target value $y$ of a sample $x$ is given by the target model $f(x) = 3-sin(3x)/x$. The target value is then perturbed by Gaussian noise, i.e., $\hat{y} = y + \epsilon = f(x) + \epsilon$, where $\epsilon \sim \mathscr{N}(0, 1.2)$.

A 10-degree polynomial function is used and trained by ridge regression, i.e., $\hat{f}(x) \sim \mathscr{O}(10)$. The hyper-parameter $\lambda$ in ridge regression is searched in $\lbrace e^{-7}, e^{-6}, e^{-5}, e^{-4}, e^{-3}, e^{-2}, e^{-1}, e^{0},  e^{1}\rbrace$. Under different values of $\lambda$, the complexities of fitting models differ accordingly. For each value of $\lambda$, 40 fitting models are learned using different random training sets. Each training set is composed by 200 samples randomly sampled from the 4000 realizations. An additional test set is constructed in the same way as training sets and is of the same size.
$\hfill\qedsymbol$
\end{example}

Let $\hat{\textbf{w}}_t = (\hat{w}_{t,1}, \cdots, \hat{w}_{t,10})^T$ be the model parameter learnt on a training set $T_t$. The model complexity of a learnt model $f_{\hat{\textbf{w}}_t}$ parameterized by $\hat{\textbf{w}}_t$ is calculated as follows: 
\begin{equation*}
    m(f_{\hat{\textbf{w}}_t }) = \sum_{i = 1}^{10} (\dfrac{i}{10} \hat{w}_{t,i})^2,
\end{equation*}
and the model complexity expectation is calculated as:
\begin{equation}\label{22}
    c(\hat{\textbf{w}}(\lambda)) =\dfrac{1}{40} \sum_{t = 1}^{40}  m(f_{\hat{\textbf{w}}_t}) = \dfrac{1}{40} \sum_{t = 1}^{40} [\sum_{i =1}^{10} (\dfrac{i}{10} \hat{w}_{t,i})^2]
\end{equation}
Details of the calculation are presented in Appendix A.

\indent

The bias, variance, and generalization error curves of Example 1 are given in Fig.~\ref{fig15}. Assumption \ref{assump1} holds with regard to this example. A clear bias-variance trade-off is presented. The bias curve decreases with respect to the employed model complexity while the variance term increases. The generalization error first decreases to its minimum, and then increases. The minimum of the average generalization error over all samples is $0.018$ and is attained around the intersection of the bias curve and the variance curve. 
\indent
\\
\indent

\begin{figure}[t]
\centerline{\includegraphics[width=0.5\linewidth]{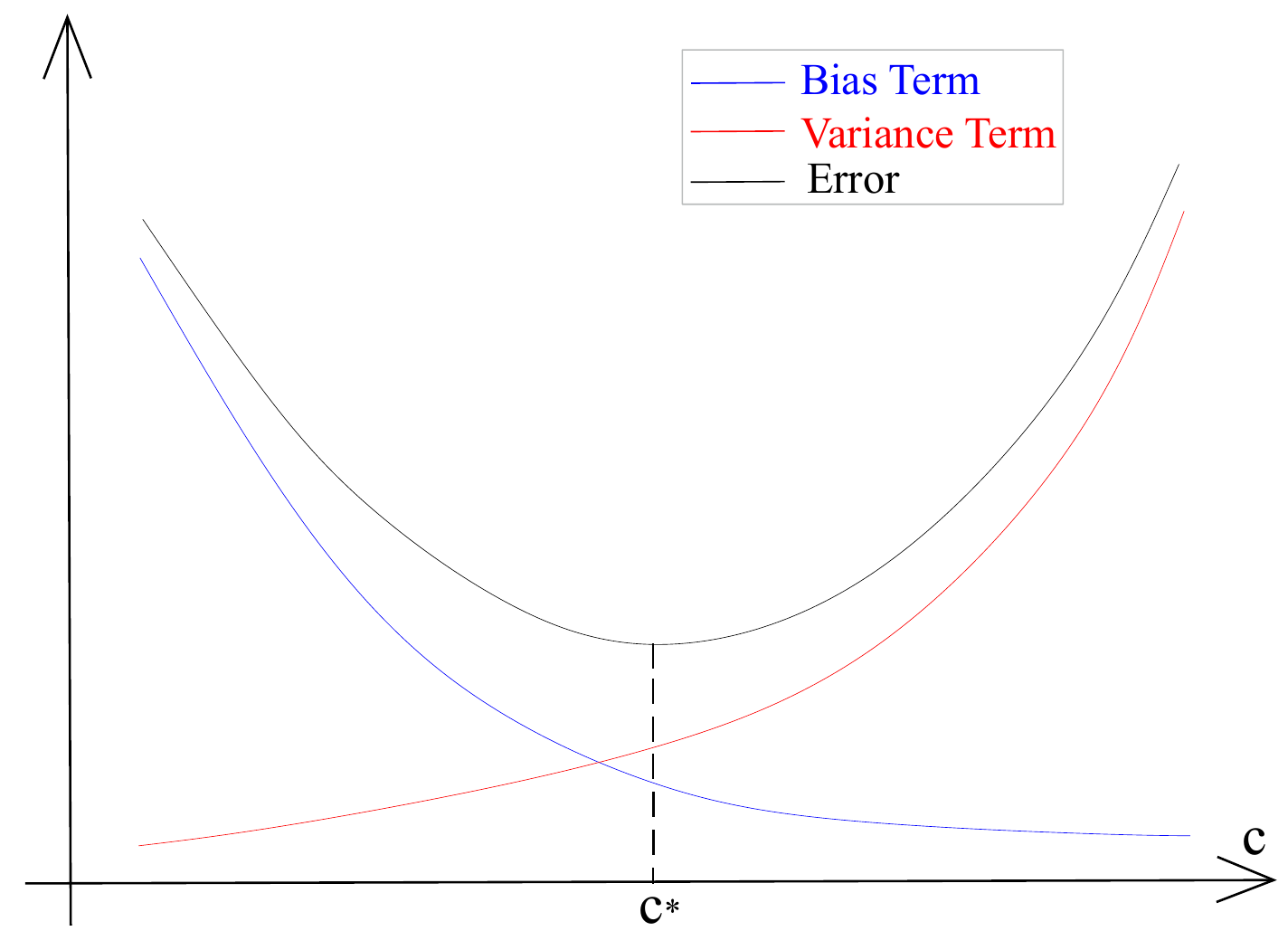}}
\caption{The bias-variance trade-off curve.}
\label{fig1}
\end{figure}

According to Assumption \ref{assump1}, the minimum generalization error is achieved when the partial derivatives of generalization error with respect to $c$ equals to zero, i.e., the sum of the partial derivatives of its bias term and the corresponding variance term with respect to $c$ equals to zero. A diagrammatic drawing of the bias-variance curve is shown in Fig. \ref{fig1}~\cite{b49,b50}. In the rest of the paper, the model complexity expectation is briefly termed as ``model complexity".

The optimal $\lambda_{h}$ and $c$ are obtained with 
\begin{equation}\label{8}
    \lambda^*_{h} = arg \min_{\lambda_{h}} Err(\lambda_{h})
\end{equation}
\begin{equation}\label{9}
    c^* = g(\lambda^*_{hyper}).
\end{equation}

\subsection{Definition of Learning Difficulty}
\indent

Eq.~(\ref{1}) is deﬁned on the whole space $\Omega$. $P(x,y)$ is often unknown. According to Eq.~(\ref{1}) we deﬁne the generalization error for a sample as follows:

\begin{definition}
    Generalization error for $x$ with label $y$:
    \begin{equation}\label{10}
        Err(x,\lambda_{h}) = \mathbb{E}_T[l(y,f(x;T,\lambda_{h}))] ,
    \end{equation}
where $l(\cdot ,\cdot)$ measures the error between the label and a prediction.
\end{definition}

Accordingly, 
    \begin{equation}\label{12}
        BiasT(x,\lambda_{h}) = l(y,\overline{f}(x;\lambda_{h}))
    \end{equation}
    \begin{equation}\label{13}
        VarT(x,\lambda_{h}) = \mathbb{E}_T[l(\overline{f}(x;\lambda_{h}),f(x;T,\lambda_{h}))]
    \end{equation}

Similar to Assumption 1, we have the following assumption:
\begin{assumption}\label{assump2}
Assume that the basic model is given and ﬁxed. The bias term for $x$ is a decreasing function of $c$, whereas the variance term for $x$ is an increasing function of $c$.   The generalization error for $x$ decreases first and then increases.
\end{assumption}

An example of non-linear regression learning is still utilized to empirically support Assumption \ref{assump2}.

\begin{example}
The target model and the calculation of model complexity in Example 1 are still used. The simulation of samples is same as described in Example 1 except the sampling strategy for $x$. Imbalance sampling is applied and each training set is consists of 100 random samples with $x \in [0,1.5)$, 50 random samples with $x \in [1.5,3.5)$, and 25 random samples with $x \in [3.5,5]$. The samples in the additional data set are sampled with the same imbalance strategy. Under each value of $\lambda$, 40 models are learned using different training sets sampling from the initial 4000 realizations.
$\hfill\qedsymbol$
\end{example}

\begin{figure}
\centerline{\includegraphics[scale=1.1]{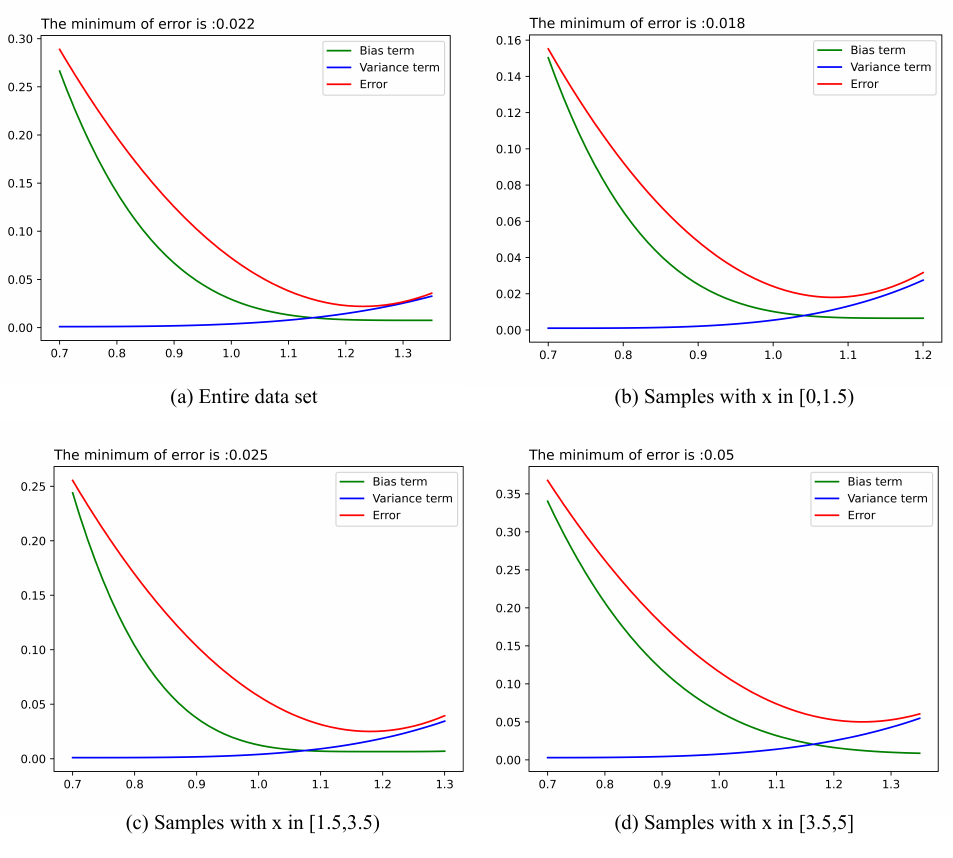}}
\caption{Illustrations of bias-variance trade-off under imbalance sampling.}\vspace{-0.1in}
\label{fig16}
\end{figure}

\begin{figure}
\centerline{\includegraphics[scale=1.1]{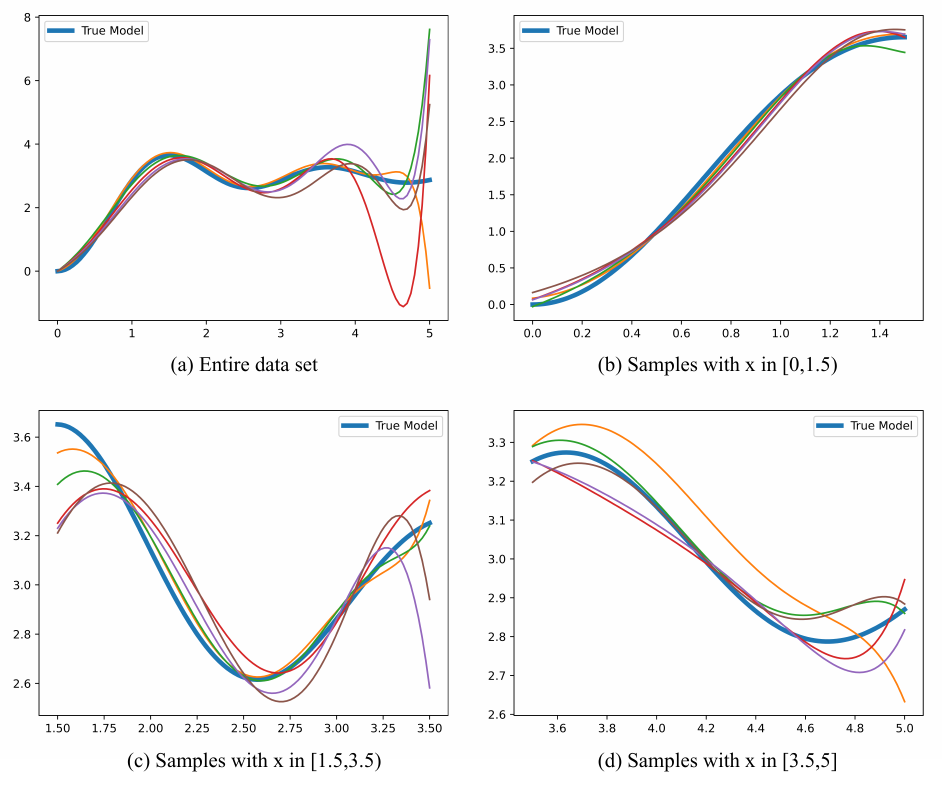}}
\caption{Illustrations of comparisons between learned and true models under imbalance sampling.}\vspace{-0.1in}
\label{fig17}
\end{figure}

The imbalance sampling aims to generate three regions comprising easy, medium, and hard samples, respectively. Samples with $x \in [0,1.5)$ are relatively easy and those with $x \in [3.5,5]$ are relatively hard. The learning curves are shown in Figs.~\ref{fig16} and \ref{fig17}. Fig.~\ref{fig16} shows the bias-variance trade-off curves for the entire data set (Fig.~\ref{fig16}(a)), the samples from $[0,1.5)$ (Fig.~\ref{fig16}(b)), the samples from $[1.5,3.5)$ (Fig.~\ref{fig16}(c)), and the samples from $[3.5,5]$ (Fig.~\ref{fig16}(d)), respectively. Under all cases, the bias term decreases with respect to the employed model complexity, while the variance term increases. The generalization error firstly decreases to its minimum, and then increases. The minimum values of generalization error varies: 0.022 for the entire data set (Fig.~\ref{fig16}(a)), 0.018 for the majority sampling part ($[0,1.5)$) (Fig.~\ref{fig16}(b)), 0.025 for the medium sampling part ($[1.5,3.5)$) (Fig.~\ref{fig16}(c)), and 0.05 for the minority sampling part ($[3.5,5]$) (Fig.~\ref{fig16}(d)). Comparing the minimum of generalization errors of three cases above(Fig.~\ref{fig16}(b-d)), the learned models perform the worst in the hard region ($[3.5,5]$) shown in Fig.~\ref{fig17}(d), as the learned models under various $\lambda$ can hardly match the target model. The learned models perform the best in the easy region shown in Fig.~\ref{fig17}(b), as the learned models basically coincide with the target model. Alternatively, both Assumptions \ref{assump1} and \ref{assump2} hold in this example.
\indent


According to Assumption \ref{assump2}, the minimum of $ Err(x,\lambda_{h})$, denoted as $Err^*_x$, is also attained when the partial derivatives of generalization error for $x$ on $c$ equals to zero, i.e., the sum of the partial deviation of the bias term and the variance term for $x$ on $c$ is zero. The corresponding value of $c$ for sample $x$ is denoted by $c_x^*$.

Accordingly, we define a theoretical measure for the learning difficulty.

\begin{definition}
    Given a fixed basic learner\footnote{Essentially, when $Err(\lambda_{h} )= \mathbb{E}_f \mathbb{E}_{(x,y)\in \Omega} \mathbb{E}_T [ \|y-f(x;T,\lambda_{h} )\|_2^2]$ is used, the learning difficulty is independent of the basic learner $f$.} $f$, the theoretical learning difficulty for a sample $x$ is
    \begin{equation}
        \begin{aligned}
            & \mathcal{LD}(x) = c_x^*=g(\lambda_{h}^*)\\
       s.t.,\ & \lambda_{h}^* = arg\min_{\lambda_{h}} Err(x,\lambda_{h}).
        \end{aligned}
    \end{equation}
\end{definition}
The relative learning difficulty between two samples $x_1$ and $x_2$ is obtained according to Definition 2. If $\mathcal{LD}(x_1 )>\mathcal{LD}(x_2)$, then $x_1$ is more difficult than $x_2$, and vice versa.

We also define a learning difficulty coefficient as follows:
\begin{definition}
    Given the optimal model complexity $c^*$ on the whole space $\Omega$ and the learning difficulty of the sample $x$, the learning difficulty coefficient ($\mathcal{LDC}$) is defined as
    \begin{equation}
        \mathcal{LDC}(x) = \dfrac{\mathcal{LD}(x)}{c^*} = \dfrac{c^*_x}{c^*}
    \end{equation}
\end{definition}
\indent
The larger the value of $\mathcal{LDC}$ is, the more difficult the sample $x$ will be. The succeeding subsection will define the easy and hard samples based on $\mathcal{LDC}$.


\subsection{Definitions of Easy and Hard Samples}
\indent

Many existing studies are based on the two or three splits for training samples, namely easy/hard and easy/medium/hard, respectively. With $\mathcal{LDC}$, the dichotomy is defined as follows:
\begin{definition}
    Given a sample $x$ and its learning difficulty coefficient $\mathcal{LDC}$, if $\mathcal{LDC}(x) \leq 1$, then $x$ is an easy sample; if $\mathcal{LDC}(x)>1$, then $x$ is a hard sample. 
\end{definition}
Definition 4 is flexible because the threshold can be a parameter instead of a fixed value. Let $\tau$ be the threshold. If $\mathcal{LDC}(x) \leq \tau$, then $x$ is an easy sample; if $\mathcal{LDC}(x)> \tau$, then $x$ is a hard sample. 
\indent

In the trichotomy, distinguishing between easy and medium or medium and hard is difficult. Accordingly, we propose the following definition for these partitions:
\begin{definition}
    Given a sample $x$ and its learning difficulty coefficient $\mathcal{LDC}(x)$, let  $\tau_e$ and $\tau_h$ be two positive parameters and $0<\tau_e<1<\tau_h$. If $\mathcal{LDC}(x) \leq \tau_e$, then $x$ is an easy sample; if $\tau_e<\mathcal{LDC}(x) \leq \tau_h$, then x is a medium sample; if $\tau_h<\mathcal{LDC}(x)$, then $x$ is a hard sample.
\end{definition}

\begin{figure}[t]
\centerline{\includegraphics[scale=0.95]{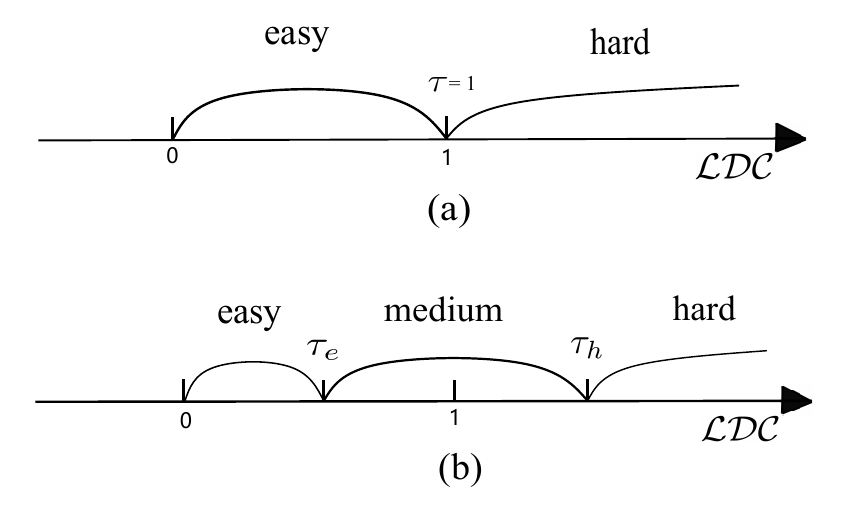}}
\vspace{-0.15in}
\caption{ Illustrations for dichotomy (a) and trichotomy (b) of samples.}\vspace{-0.15in}
\label{fig2}
\end{figure}

\indent
The two parameters depend on the concrete application tasks and data characteristics. The above two definitions describe the dichotomy and trichotomy for samples as shown in Fig. \ref{fig2}. Some samples are quite hard and are harmful to learning process. We can also define quite-hard samples if $\mathcal{LDC}(x)>\tau_q$ and $\tau_q > \tau_h$. 

Our definition for easy/medium/hard samples is consistent with the descriptive definition given by Arpit et al.~~\cite{b74}, as the model complexity will increase with the increasing of training epoch gradually. Some other studies also hold the similar view. For example, Charrerjee and Zielinskix~\cite{b75} observed that easy ImageNet samples are learned early and hard ImageNet samples are learned late. 
 
This section mathematically defines learning difficulty of samples as well as easy/medium/hard samples. Unfortunately, these mathematical definitions are not ready for algorithmic implementation due to lack of a consensus calculation for model complexity (the function $m(\cdot)$ in Eq.~\ref{7}). There are two strategies to measure the learning difficulty with the illumination of our definitions. The first is to seek an existing measure that best suits our definitions; the second is to derive a new measure on the basis of the definitions. This study chooses the second path.
Results in Example 2 indicate a positive correlation between optimal model complexity and minimum generalization error. Moreover, the definition of optimal model complexity is closely related to the generalization error. As a consequence, we propose a new measure for learning difficulty with generalization error.

\section{The Proposed Measure for Learning Difficulty}
\indent

The learning difficulties of samples depend on various factors\footnote{This study does not consider the factors from the employed learning models.}. Each of existing measures focuses on one or partial factors. This section first summaries the main factors that determine the learning difficulty of a sample. Then, the connection between the model complexity and generalization error is analyzed based on the summarized influential factors. Finally, the concrete measure is introduced and the algorithmic steps are presented.

\subsection{Influential Factors for Learning Difficulty}
\indent

So far, there has been no study that comprehensively summarizes the factors that determine the learning difficulty of a sample. In machine learning, differences among samples mainly lie in noise level, spacial location, neighbourhood, and overall distribution. Illuminated by these differences and the heuristic inspirations considered in previous measures for learning difficulty, the main influential factors are roughly summarized as follows:
\begin{itemize}
    \item Data quality. Both feature and label noises affect the learning difficulty of samples. Young et al.~\cite{b63} found that high signal-to-noise ratio signifies high feature noise level and generates low data quality, which hinders the optimization of the learning task and is harder to be well learnt. Su et al.~~\cite{b64} revealed that mislabeled images are low-quality data for the learning task and are of high difficulty.  
 
    \item Sample margin. The sample margin is defined as the distance between the sample and the true decision boundary. Huang and Yang~\cite{b65} considered that samples with small margins are hard to learn.
    \item Uncertainty. The (model) uncertainty of a sample is usually measured by the information entropy of its prediction~\cite{b66}. The higher the information entropy is, the ampler the information is contained by the sample. Samples with higher uncertainty are more difficult to adequately learn~\cite{b67}.
    \item Category distribution. Oinar et al.~\cite{b68} showed that category with fewer samples, which is also called tail category, is usually more challenging than category with more samples, which is known as head category.
\end{itemize}

The learning difficulty of a specific sample is usually determined simultaneously by two or more factors. A measure that only considers partial factors will perform poorly when the application scenario changes.

\subsection{Connection Between Model Complexity and Generalization Error}
\indent

To our knowledge, there is no study that directly investigates the connection between model complexity and generalization error for a sample\footnote{Perez and Louis~\cite{b76} observed a linear relation between model complexity and the overall generalization error (in Fig.~1 in their paper) instead of that for each sample.}. Nevertheless, a positive correlation can be approximately established for them bridging by the four influential factors.

First, previous studies explicitly mention a positive correlation between the aforementioned influential factors and the optimal model complexity (defined as learning difficulty of a sample in this study).  1) Noisy samples result in a higher model complexity if these samples are well-learned \cite{b51,b52}. 2) To correctly learn a sample in tail categories, strategies such as over-sampling is usually introduced and leads to a higher model complexity \cite{b53,b54}. 3) To well learn the sorts of small-margin samples, the learned boundary turns to be more complex \cite{b55,b56}. 4) A more uncertain sample has a larger variance of prediction. A more complex model is needed to reduce its variance, which is to correctly classifying \cite{b57,b58,b59}.

Second, the positive relationship between the influential factors and the generalization error is also discussed in previous literature\footnote{In fact, existing studies focus on the generalization error over the whole data space rather than a local region or a single sample. Nevertheless, the positive correlation between factors and $Err^*_x$ of a single sample is theoretically verified in our continuous study. The theoretical proofs are uploaded to Github source repository.}. In terms of data quality, Wang et al.~\cite{b69} 
revealed
that negative impacts of noisy implicit feedback occurs to the minimization of generalization error. Castells et al.~\cite{b70} concluded that noisy samples tend to be harder and injure the model generalization. In terms of sample margin, Zhang et al.~\cite{b5} concluded that samples close to boundary, are further from reaching near-zero generalization error than samples away from boundary. In terms of uncertainty, Pagliardini et al.~\cite{b72} improved the model's generalization by estimating uncertainty quantification and perturbing high uncertain samples. In terms of category distribution, Gautheron et al.~\cite{b73} derived a bound of generalization error in metric learning by involving the proportion of minority examples who throw higher generalization error values.

Based on above-mentioned analysis, an approximately positive correlation between generalization error ($Err^*_x$) and model complexity ($c^*_x$) is drawn and motivates a practical measure of learning difficulty. Indeed, Fig.~\ref{fig16} also demonstrates a direct and positive correlation between $Err^*_x$ and $c^*_x$. A higher $Err^*_x$ corresponds to a larger $c^*_x$. A solid proof under typical cases is left as our future work.

\subsection{The Proposed Measure}

\indent

Under the discussion in Section 4.2, an approximate approach is proposed based on $Err_{x}^*$, and is utilized as the practical measure of learning difficulty. Most existing measurement methods utilize the training loss (or loss variance) which can be considered as an approximation of the bias (or the variance) term of $Err_{x}^*$ as a measurement of learning difficulty. Nevertheless, few studies consider the bias and the variance terms simultaneously. 
\indent

Considering that it is also infeasible to calculate $Err_{x}^*$ traversing all values of $\lambda_h$, we only calculate the generalization errors $Err(x, \lambda_h)$ for each sample with the same\footnote{In the experiments, there are slight differences among different samples as the optimal epochs are not identical in different training runs.} reasonable $\lambda_h$ to approximate the learning difficulty. Specifically, the proposed approach adopts the cross-validation strategy to calculate the average learning errors for each sample. First, the whole training set is divided into $M$ folds. $M-1$ folds are alternatively used for training, and the trained model is used to predict the label of all training samples. This cross-validation process is repeated for $K$ times. Each sample receives $K*M$ predictions, with which can we calculate the average prediction of each sample. Second, average losses and variance of losses for each training sample are calculated using corresponding average predictions.

Let $p^k_{i,m}$ be the prediction of $x_i$ in the $m^{th}$ cross-validation of the $k^{th}$ repeat. Then, according to~\cite{b41}, we calculate:
\begin{equation}\label{26}
    \overline{p}_i = exp\lbrace \dfrac{1}{M*K} \sum_{m,k} log(p^k_{i,m}) \rbrace.
\end{equation}
Subsequently, the bias and the variance terms are calculated as follows
\begin{equation}\label{27}
Bias_i \approx l_{CE}(y_i, \overline{p}_i) ,
\end{equation}
\begin{equation}\label{28}
    Var_i \approx \dfrac{1}{M*K} \sum_{m,k} l_{CE}(\overline{p}_i, p^k_{i,m}),
\end{equation}
where $l_{CE}$ is the standard cross-entropy loss. The actual value of learning difficulty of $x_i$ is
\begin{equation}\label{29}
    Err(x_i, \lambda_h) \approx Bias_i + \mu Var_i\ ,
\end{equation}
where $\mu$ is a tuning factor for the variance.
The value of $Err(x_i,\lambda_h)$ is used as the learning difficulty for $x_i$. This approach is called generalization error-based learning difficulty (GELD) measurement. The detailed steps of GELD are shown in Algorithm \ref{algo1}. The primary difference between our approach and the existing loss-based/cross-validation-based methods lies in that our approach does not discard the variance term but combines the importance of both term of generalization error. If $\mu = 0$, then GELD is similar to the conventional cross-validation-based methods. Several existing methods including O2UNet~\cite{b17} also point out that hard samples have high loss variances.

\renewcommand{\algorithmicrequire}{\textbf{Input:}}
\renewcommand{\algorithmicensure}{\textbf{Output:}}
\begin{algorithm}
\caption{GELD}\label{algo1}
\begin{algorithmic}[1]
\Require $T = \lbrace x_i \rbrace_{i=1}^N=\lbrace (x_i,y_i)\rbrace_{i=1}^N$, validation data, $M$, $K$, $\mu$, and $\lambda_{h}$.
\Ensure $ Err(x_i, \lambda_h)$, $i = 1, \cdots, N$.\\
\FOR $k\leftarrow 1$ \TO $K$:
\State \indent Randomly split $T$ into $T_1^{(k)}, \cdots,\ T_M^{(k)}$;
\State \indent \FOR $m\leftarrow 1$ \TO $M$:
\State \indent \indent \text{Perform the training on $T-T_m^{(k)}$;}
\State \indent \indent \text{Select the model with the validation data;}
\State \indent \indent \text{Predict $p^k_{i,m}$ of each $x_i$.}

\State Calculate $\overline{p}_i$ using Eq. (\ref{26}) for each $x_i$.
\State \text{Calculate $Bias_i$ and $Var_i$ using Eqs. (\ref{27}) and (\ref{28}).}
\State \text{Calculate $ Err(x_i, \lambda_h)$ using Eq. (\ref{29}).}
\end{algorithmic}
\end{algorithm}
\bigskip

\section{Theoretical Analysis Based on Theoretical Definitions}
\indent

Definitions 1-5 theoretically describe our learning difficulty measure and clarify easy, medium, and hard samples. This section conducts a theoretical analysis for the weighting strategies to further illuminate the rationality and intrinsic value of our theorem, given that the weighting strategies in machine learning are mainly based on learning difficulties, such as Adaboost~\cite{b31}, SPL~\cite{b3,b15}, and Focal loss~\cite{b16}. Under our definition of learning difficulty, above-mentioned weighting strategies can be better rationalised and further comprehended. First, the weighted generalization error\footnote{When generalization error is defined over the entire sample space, regard $Err(\lambda_h)$ as $Err(\Omega , \lambda_h)$ for simplicity.} is defined as follows:
\begin{equation}
\label{16}
    \begin{aligned}
      Err^{\text{w}}(\lambda_{h}) &= \sum_{y \in \Omega_Y}   P(y) \int_{x \in \Omega_X} \omega (x)Err(x,\lambda_{h}) p(x\vert y)\mathrm{d}x \\
      & = \sum_{y \in \Omega_Y} P(y) \int_{x \in \Omega_X} \omega (x)BiasT(x,\lambda_{h})p(x\vert y)\mathrm{d}x \\
      & + \sum_{y \in \Omega_Y} P(y) \int_{x \in \Omega_X} \omega (x)VarT(x,\lambda_{h})p(x\vert y)\mathrm{d}x \\
      & + \delta^{'}_{e}\ ,
    \end{aligned}
\end{equation}
where the non-negative weighting function $\omega (x)$ is defined over the entire sample space $\Omega$, and $\delta^{'}_{e}$ is the irreducible noise. 

\subsection{Propositions}
\indent

Both the bias and the variance terms will change when the weighting strategy is used on samples as shown in Eq.~(\ref{16}).
Thus, a special case is first analyzed.

\begin{proposition}
 Assume that $BiasT(\lambda_{h})$ and $VarT(\lambda_{h})$ change. Let $\overline{BiasT}(\lambda_{h})$,  $\overline{VarT}(\lambda_{h})$, and $\overline{Err}(\lambda_{h})$ be the new bias, variance terms, and the generalization error, respectively. If the partial deviation of $\overline{Err}(\lambda_{h})$ on the current $c^*$ is negative, then the new optimal model complexity $\overline{c}^*$ will be larger then $c^*$, and vice the versa.
\end{proposition}
The proof is contained in Section B.1. in Appendix.
\begin{proposition}
 If $\omega(x)$ in Eq.~(\ref{16}) is a constant value, then $c^*$ remains unchanged.
\end{proposition}
The proof is simple and omitted.
\begin{proposition}
 Consider a sample region $\Omega^r \subset \Omega$ in which the value of $\mathcal{LDC}$ for each sample in $\Omega^r$ is larger than one.  If a constant weight $\omega$ larger than one is placed on each sample of $\Omega^r$ and the weights of other samples in $\Omega$ remain one, then the new optimal model complexity will become larger.
\end{proposition}
\indent
The proof is contained in Section B.2. in Appendix.

\indent
This proposition is in accordance with the idea that when the weights of hard samples are increased, the learned model will become more complex than the original model. Based on this proposition, we could extend a corollary more generalized.
\begin{corollary}
Consider a sample region $\Omega^r \subset \Omega$ whose learning difficulty coefficient $\mathcal{LDC}$ for each sample in $\Omega^r$ is larger than one. If a weight larger than the original weight is placed on each sample in $\Omega^r$, and the weights on other regions remain unchanged, the new optimal complexity will become larger.
\end{corollary}
\indent
The proof is contained in Section B.3. in Appendix.
\begin{proposition}

 Consider a sample region $\Omega^r \subset \Omega$ whose value of $\mathcal{LDC}$ for each sample in $\Omega^r$ is smaller than one. If a constant weight larger than one is placed on each sample of $\Omega^r$, and the weights of other samples in $\Omega$ remain one, then the new optimal model complexity $c'^*$ will become smaller.
\end{proposition}

The proof is similar to that for Proposition 3 and Corollary 1 and omitted. In numerous weighting strategies, the rationale is to modify the contributions of easy, medium, and hard samples. Therefore, the following propositions are presented.
 \begin{proposition}
  Assuming that the original weight of each sample is $\omega_0(x)$. Let $\omega(x)=u(\mathcal{LD}(x))$ be a new weighting function for a sample $x$. If u is non-decreasing and satisfies that $0 \leq \min \omega(x) < \max \omega(x)$, then the new optimal complexity is larger than the original optimal complexity.
 \end{proposition}
 \indent
The proof is contained in Section B.4. in Appendix.

 \begin{corollary}
  Let $\Omega^e$, $\Omega^m$, and $\Omega^h$ be a trichotomy for the whole space $\Omega$, and they represent the regions for easy, medium, and hard samples, respectively. Let $\omega(\cdot,\cdot)$ be a region weighting function over the three data regions, and assume that the weights in each region are identical for each sample. Note $x \in \Omega^e; x' \in \Omega^m; x'' \in \Omega^h$. If $\omega(x) \leq \omega(x') \leq \omega(x'')$ and $\omega(x)< \omega(x'')$ hold, then the new optimal complexity will become larger.
 \end{corollary}
 The proof is simple and omitted.
\begin{proposition}
  Assume that the original weight of each sample is $\omega^0(x)$. Let  $\omega(x)=u(\mathcal{LD}(x))$ be a weighting function for a sample $x$. If u is non-increasing and satisfies that $ 0 \leq \min \omega(x) < \max \omega(x) $, then the new optimal complexity is smaller than the original optimal complexity.
\end{proposition}
\begin{corollary}
 Let $\Omega^e$, $\Omega^m$, and $\Omega^h$ be a trichotomy for the whole region $\Omega$ and they are the regions for easy, medium, and hard samples, respectively. Let $\omega(\cdot,\cdot)$ be a region weighting function over the three data regions, and assume that the weights in each region are identical for each sample. Note $x \in \Omega^e; x' \in \Omega^m; x'' \in \Omega^h$. If $\omega(x) \geq \omega(x') \geq \omega(x'')$ and $\omega(x) > \omega(x'')$ hold, then the new optimal complexity will become smaller.
\end{corollary}
\indent
Propositions 1-6 and the associated corollaries are about the weighting on generalization errors and also the losses. They establish a theoretical framework for the analysis of the learning difficulty-aware weighting strategies in learning.

\subsection{Explanations for Several Classical Methods}
\indent

We rationalises several typical learning methods which assign weights on samples based on learning difficulties\footnote{It should be noted that the difficulty measures in these methods are not equal to our proposed theoretical measure. Nevertheless, we assume that their employed measures are in accordance with ours in their contexts to facilitate further theoretical investigation.}.

\subsubsection{Adaboost}
\indent

Adaboost is a classical ensemble learning algorithm. In each epoch, it learns a new model based on the updated weights on samples defined as follows:
\begin{equation}\label{17}
    \omega_i^t = \dfrac{\omega_i^{t-1}}{z^{t-1}}exp(-\alpha y_i f^{t-1}(x_i)),
\end{equation}
where $\omega_i^t$ is the weight in the $t^{th}$ epoch, $z^{t-1}$ is a normalized factor, $f^{t-1}$ is the learned weak classifier in the $(t-1)^{th}$ epoch, and $\alpha$ is a positive weight for $f^{t-1}$. According to Eq.~(\ref{17}), if $x_i$ is mis-predicted by $f^{t-1}$, then the weight of $x_i$ will become larger in the next epoch. If $x_i$ is correctly predicted by $f^{t-1}$, then the weight of $x_i$ will become smaller in the next epoch. In essence\footnote{Indeed, the $\mathcal{LDC}(x_i)$ can be seen as being approximated by $exp(y_if(x_i))$ in Adaboost.}, the weight in Eq.~(\ref{17}) can be written as follows:
\begin{equation}
    \omega_i^t = \omega_i^{t-1} u(\mathcal{{LD}}(x_i)),
\end{equation}
where
\begin{equation}
    u(\mathcal{LD}(x_i)) =\left\{
    \begin{array}{rcl}
    \dfrac{exp(\alpha)}{z^{t-1}}       &      & \text{if}\ \mathcal{LDC}(x_i) > 1\\
    \ \\
    \dfrac{exp(-\alpha)}{z^{t-1}}      &      & \text{otherwise}
    \end{array} \right.
\end{equation}

Obviously, $u(\mathcal{LD}(x_i))$ is an increasing function over learning difficulty. According to Proposition 5, the new model complexity becomes larger than the original one. Specially, the learned new classifier $f^{t}$ is more complex than that in the $(t-1)^{th}$ epoch if the learner is not as simple as the decision trump. Therefore, the new classifier and the whole ensemble classifier become more complex with the increase in epoch. 

Two aspects determine the complexity of the final ensemble model greatly:
\begin{itemize}
    \item Power of the basic model. If the basic model has strong classifier, such as SVM, the learned model will become highly complex with the increase in epoch and overfitting is inevitable. A weak learner can avoid this situation.
    \item Number of maximum epochs. If the maximum epoch is large, then the ensemble model in the last few epochs will become highly complex when noises exist. Accordingly, overfitting may occur.
\end{itemize}
\indent

A natural improvement is that high weights above a threshold are restricted. This condition makes the model less complex. A famous modification with solid theoretical basis is soft margin boosting~\cite{b12}. The weight is calculated as follows:
\begin{equation}\label{20}
    \omega_i^t = \dfrac{\omega_i^{t-1}}{z^{t-1}}exp(-\alpha y_i f^{t-1}(x_i) - C \zeta_i^{t-1} \vert b^{t-1}\vert),
\end{equation}
where $C(\geq 0)$ is a hyper-parameter, $\zeta_i^{t-1}$ is the average weight of the $i^{th}$ sample up until the $(t-1)^{th}$ iteration, and $b^{t-1}$ is a factor that reflects the classification performance in the $(t-1)^{th}$ iteration. If $C > 0$, then the above weight is smaller than the weight in Eq.~(\ref{17}) when the sample is often misclassified up until the $(t-1)^{th}$ iteration, and vice versa. Based on Corollary 1, on the contrary, the optimal complexity of the learned model based on the above weighting scheme will be smaller than that of the model based on Eq.~(\ref{17}).

\subsubsection{SPL}
\indent

SPL trains models from easy samples and adds hard samples with the increasing training epoch. Its objective function is as follows:
\begin{equation}\label{19}
    \min \limits_{\Theta, v_i \in \lbrace 0,1 \rbrace} \sum_i v_i l_i - \lambda v_i ,
\end{equation}
where $\Theta$ is the model parameter, $v_i$ is the sample weight, and $\lambda >0$ is a hyper-parameter and increased with the epoch. Theoretically\footnote{Indeed, the $\mathcal{LD}(x_i)$ can be seen as being approximated by $l_i$ in SPL.}, the weight in SPL is defined as follows

\begin{equation}
    \omega_i = \left\{
    \begin{array}{rcl}
    1       &      &\mathcal{LD}(x_i) \leq \lambda\\
    \ \\
    0      &      & \text{otherwise} .
    \end{array} \right. 
\end{equation}

In each new epoch, the weights of some hard samples are changed from zero to one with the increase in $\lambda$. According to the Corollary 1, the optimal model complexity will become larger. Alternatively, SPL obtains simple models in the initial epochs and gradually yields complex models.

\begin{figure}[t]
\centering
\includegraphics[width=1\linewidth]{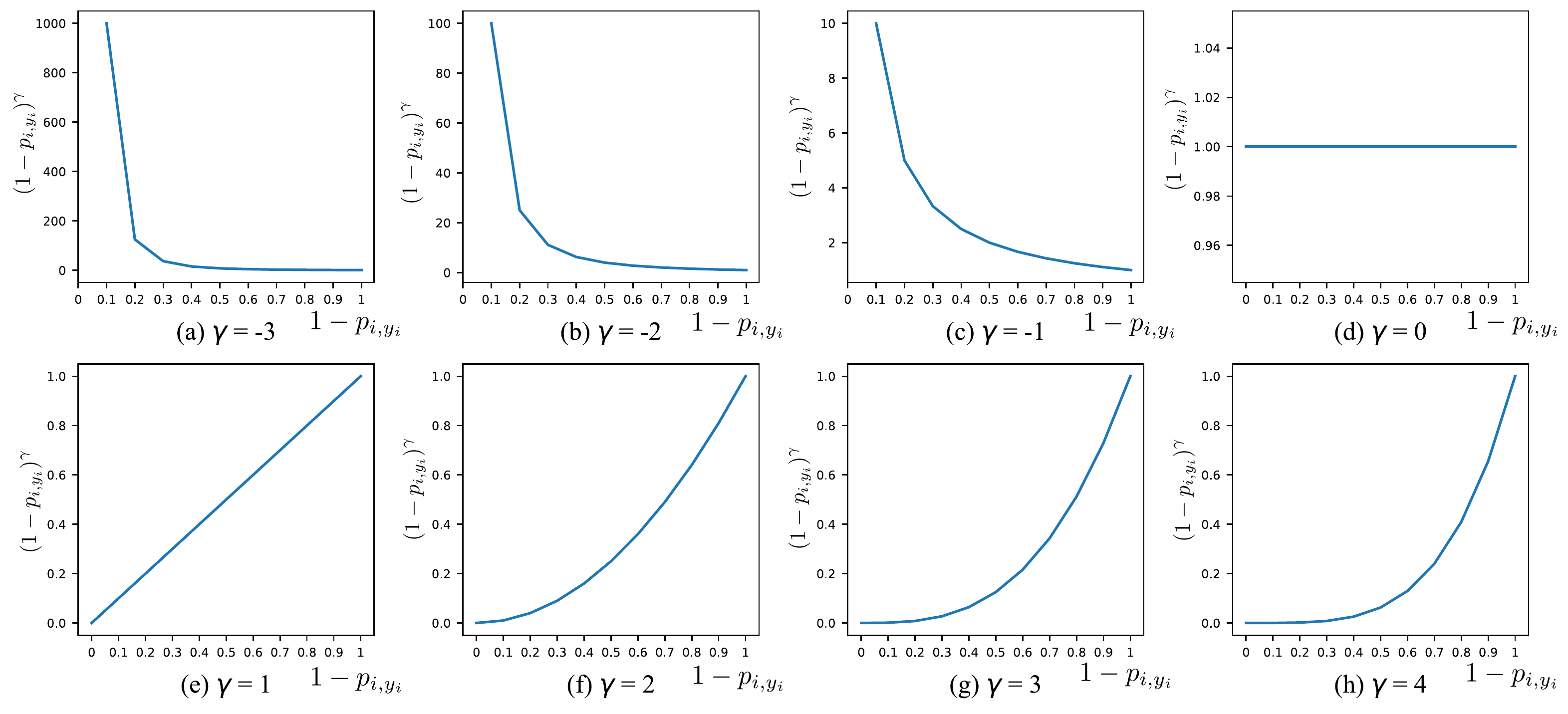} 
\caption{ Curves of Focal loss with different values of $\gamma$.}
\label{fig3}
\end{figure}

\subsubsection{Focal loss}
\indent

Focal loss assigns each sample a weight as follows:
\begin{equation}\label{23}
    \omega_i = (1-p_{i,y_i})^{\gamma},
\end{equation}
where $p_{i,y_i}$  is the estimated SoftMax value of $x_i$ on the ground-truth label in the current model, and $\gamma$ is positive. The motivation of Focal loss is to exert (relatively) larger weights on hard samples than simple ones. Focal loss utilizes the value of $1 - p_{i,y_i}$ as an indicator of learning difficulty. To better understand Focal loss, we first theoretically define a weight:
\begin{equation}\label{24}
    \omega_i =   \left( \dfrac{\mathcal{LD}(x_i)}{\max \mathcal{LD}(x_i)} \right) ^{\gamma}.
\end{equation}

\begin{figure}[htbp]
\centerline{\includegraphics[scale=0.6]{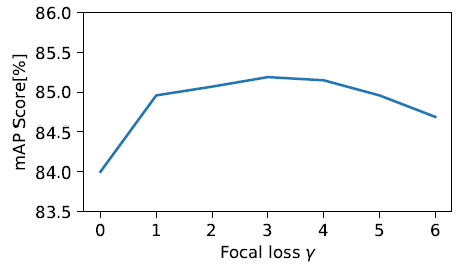}}\vspace{-0.15in}
\caption{ Detection performance with the variations of $\gamma$~\cite{b36}.}
\label{fig4}
\end{figure}

According to Proposition 5 and Corollary 2, the new optimal complexity will be increased. We further obtain the following conclusion:
\begin{corollary}
 The larger the value of $\gamma$, the larger the optimal complexity will be, i.e., $\forall \gamma_1 < \gamma_2, c^*(\gamma_1) < c^*(\gamma_2)$.
\end{corollary}
The proof is similar to that for Proposition 4. Alternately, $\gamma$ controls the model complexity. Consequently, if $\gamma$ is quite large, then the learned model will be quite complex which affects the generalization capability of the model. If $\gamma$ is smaller than zero, then the learned model will be simpler than the learned model when no weights are used (i.e., $\gamma = 0$).  Fig. \ref{fig3} shows the curves of Focal loss when $\gamma$ is searched in $\lbrace -4, -3, -2, -1, 0, 1, 2, 3, 4 \rbrace$. Corollary 4 is also supported by the empirical observations~\cite{b36} shown in Fig.  \ref{fig4}. A small(large) $\gamma$ will result in under-fitting(over-fitting).

Focal loss is actually the strategy that uses the weight in Eq.~(\ref{23}) to approximate the weight defined in Eq.~(\ref{24}):
\begin{equation}
    \left( \dfrac{\mathcal{LD}(x_i)}{\mathcal{LD}_{max}}\right) ^{\gamma} \approx (1-p_{i,y_i})^{\gamma}.
\end{equation}

\begin{remark}
The proposed propositions and corollaries in Section 5.1 are in accordance with intuitions on the model complexity variations when applying weighting for learning. The explanations for the three classical methods are also reasonable and partially supported by empirical observations as shown in Fig. \ref{fig4}. The above analysis and explanations support the rationale of our proposed learning difficulty theory.
\end{remark} 

\section{Experiments}

\indent

As previously mentioned, learning difficulty is heavily affected by the data quality, sample margin, uncertainty, and category distribution. Therefore, four different scenarios are designed to evaluate the effectiveness of the proposed practical measure.

\subsection{Measurement under Noise Detection}
\indent

Two benchmark image classification data sets~\cite{b33}, namely, CIFAR10 and CIFAR100 are used. There are 10 classes in CIFAR10 and 100 classes in CIFAR100. On both sets, there are 50,000 images for training and 10,000 images for testing. The test images are used as the validation data for CIFAR10 and CIFAR100. In this scenario, noises contain two types. The first type is noisy labels ($y$), while the second consists of noisy images ($x$). The competing methods are as follows:
\begin{itemize} 
    \item \textbf{Loss}. The losses for each sample in the epoch with the highest validation accuracy are used for measurement. 
    \item \textbf{Average loss (AveLoss)}. The average values of losses of the last 100 epochs are used for measurement. 
    \item \textbf{O2UNet~\cite{b17}}. As previously introduced, this method adopts a cyclical training procedure and the average loss of each sample in the procedure is used.
    \item \textbf{MentorNet~\cite{b22}}. This method pertaining to curriculum learning, uses the output weights of the teacher network with the highest accuracy on the validation set to present the possibility of being correct. A smaller weight indicates the sample is more difficult to learn.
    \item \textbf{Co-teaching~\cite{b21}}. Two networks are trained. For each sample, the smaller one of the two losses given by the two networks is regarded as the learning difficulty.
    \item \textbf{Variance of gradients (VOG)~\cite{b6}}. This method relies on the variances of the gradient norms of each sample cross different training epochs. A high VOG value indicates a large difficulty for a sample.
    \item \textbf{Our proposed method GELD}. The detailed steps are presented in Algorithm 1.
\end{itemize}

\begin{figure*}[htbp]
\centering
\includegraphics[scale=0.38]{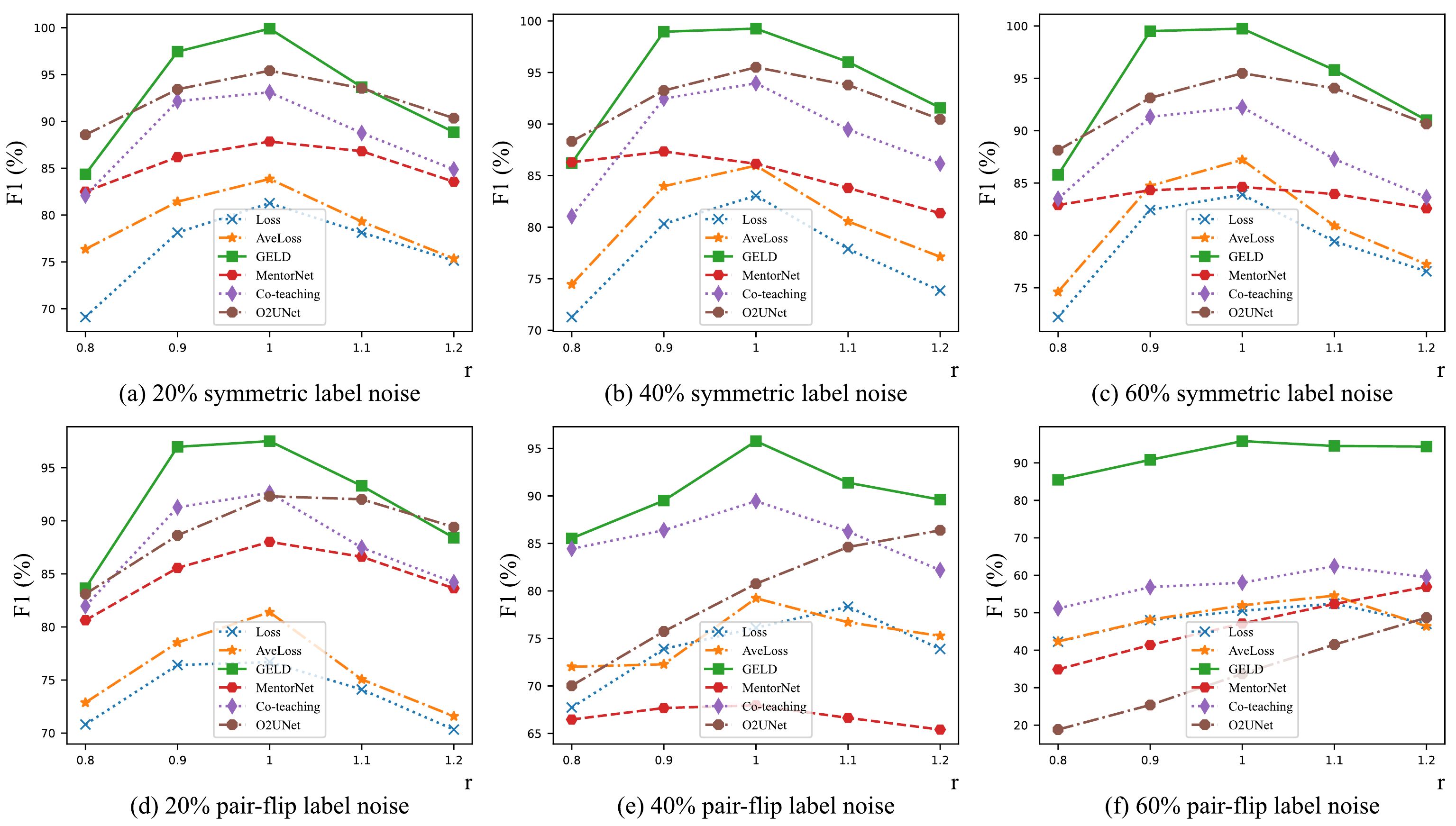}
\caption{The F1 scores ($\%$) of the competing methods on CIFAR10 under different sub-types and rates of label noises.}
\label{fig5}
\end{figure*}

In the methods of Loss, AveLoss, O2UNet, VOG, and our GELD, ResNet-34~\cite{b38} is used as the base network. The hyper-parameters of ResNet-34 used in~\cite{b38} are followed. Specifically, the batch-size is 128, the SGD optimizer has a momentum of 0.9, and the weight decay is 1e-4. The learning rate of the first 40 epochs is 0.1 and is multiplied by 0.1 for every 40 epochs. Each model is learned for 200 epochs. The default settings of MentorNet and Co-teaching are borrowed from the corresponding papers~\cite{b22,b21}. O2UNet and VOG contain specific hyper-parameters other than those of ResNet-34. These parameters follow the setting in the original paper~\cite{b6,b17}. In our GELD,
$K$ and $M$ are set as 5 and 6 respectively, for GELD, except for the part \textit{E} (the discussion of the impact of $(K,M)$ value). The tuning factor $\mu$ of GELD is set as 1.

Let $v$ be the noise rate. The result evaluation scheme used for O2UNet~\cite{b17} is followed. In each method, the top-$50000*v*r$ samples are selected as its detected noisy samples according to its estimated difficulties, where $r \in \lbrace 0.8, 0.9, 1, 1.1, 1.2\rbrace$. Then, the whole detection is repeated three times for each method and the average F1 values on the detection results are calculated and compared. A high F1 value indicates a good performance in noisy label detection and thus the learning difficulty measurement. 

\subsubsection{Noisy Label Detection}
\indent

Two sub-types of noises are used, namely, symmetric and pair-flip. The symmetric noise describes mislabeling to each other classes of equal possibility. In pair-flip, labelers may make mistakes only within very similar class. The noise rate is set as $20\%$, $40\%$, and $60\%$, respectively. The detailed noise setting in \cite{b21} is followed. 

\begin{figure*}[ht]
\centering
\includegraphics[scale=0.38]{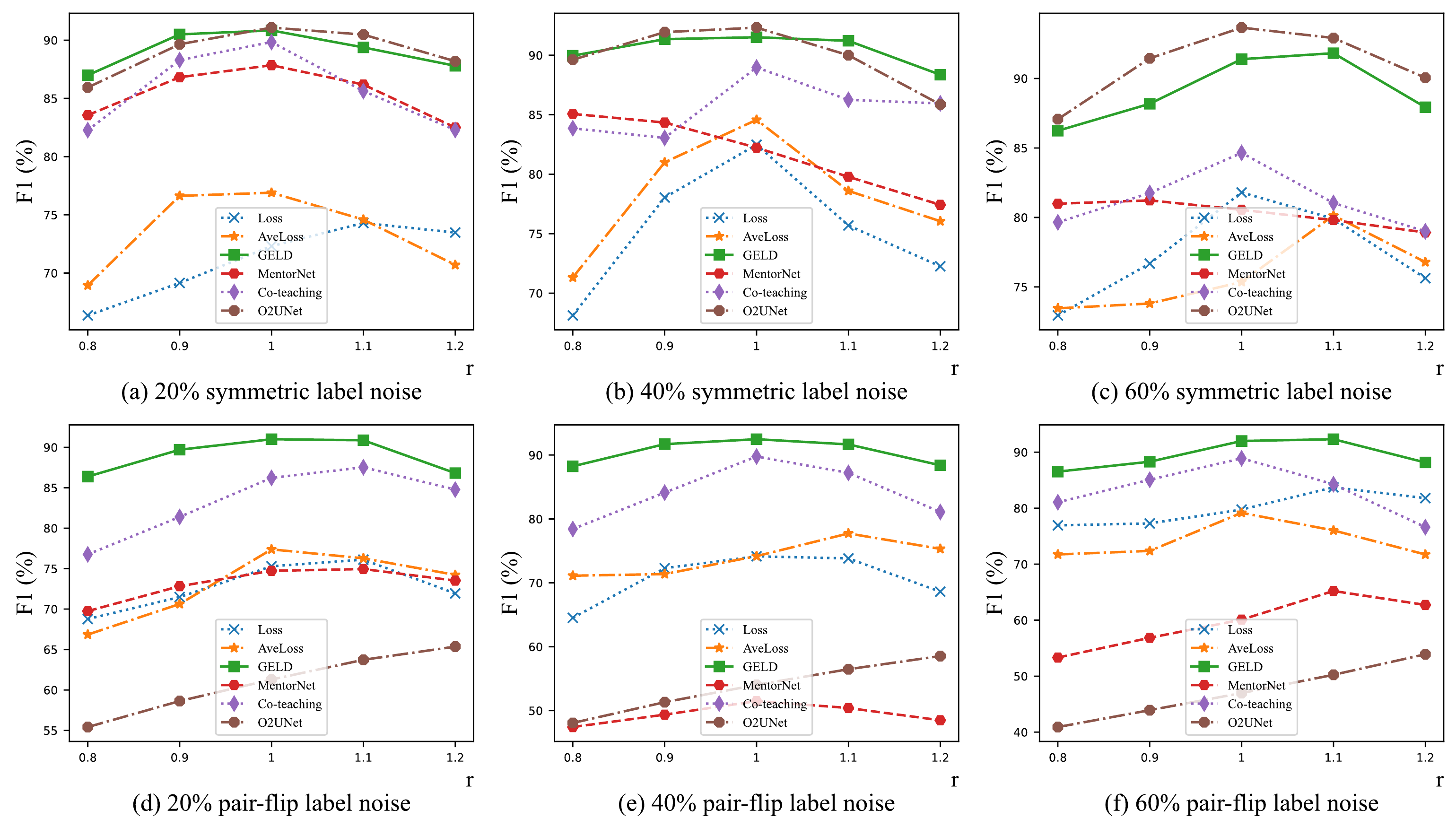}
\caption{The F1 scores ($\%$) of the competing methods on CIFAR100 under different sub-types and rates of label noises.}
\label{fig6}
\end{figure*}

 Figs. \ref{fig5} and \ref{fig6} show the detection performances of the competing methods on CIFAR10 and CIFAR100, respectively. Our proposed approach GELD achieves the highest F1 values in most cases. Although O2UNet outperforms GELD under the symmetric noise sub-type on CIFAR100 (Figs. \ref{fig6}(a), (b) and (c)), its performances are quite poor under the pair-flip noises. The performance of the widely-used method Loss is poor and it achieves the worst F1 scores in several cases. Loss is not an ideal measurement for the easy and hard samples even though it does not require additional computational cost.


 

In our approach GELD, $\mu$ can be tuned. Table \ref{tab1} shows the performance variations of GELD under the pair-flip noise sub-type and different values of $\mu$ in Eq.~(\ref{29}). When the value is larger than one, higher F1 values are achieved. These results reveal the importance of the variance term during the evaluation of the learning difficulty.

\renewcommand{\arraystretch}{1.1}
\begin{table}[h]
\begin{center}
\begin{minipage}{234pt}
\caption{$F1$ scores (\%) of GELD under various values of $\mu$.}\label{tab1}%
  \begin{tabular}{c|c|c|c|c|c|c|c}
    \toprule
   &$\mu$ & 0.5 & 0.75 & 1 & 1.25 & 1.5 & 2\\
    \midrule
     \multirow{3}*{\rotatebox{90}{CIFAR10}} &$20\%$ &   90.48 & 92.01 & 93.67 & 93.99 & 94.83& 95.06\\
    \cline{2-8} & $40\%$ & 93.95 & 93.99 & 96.03 & 96.00 & 96.00 & 95.39\\
    \cline{2-8} & $60\%$ & 91.60 & 92.47 & 95.80 & 95.67 & 95.52 & 94.99\\
  \bottomrule
    \midrule
     \multirow{3}*{\rotatebox{90}{CIFAR100}} &$20\%$ &   89.74 & 90.00 & 90.50 & 90.71 & 91.00 & 91.67\\
    \cline{2-8} & $40\%$ & 90.04 & 91.39 & 91.51& 92.00 & 92.33 & 92.97\\
    \cline{2-8} & $60\%$ & 90.33 & 91.05 & 91.39 & 91.49 & 92.41 & 93.22\\
  \bottomrule
\end{tabular}
\end{minipage}
\end{center}
\end{table}

\subsubsection{Noisy Image Detection}
\indent

\begin{figure}[htbp]
\centering
\includegraphics[width=0.85\textwidth]{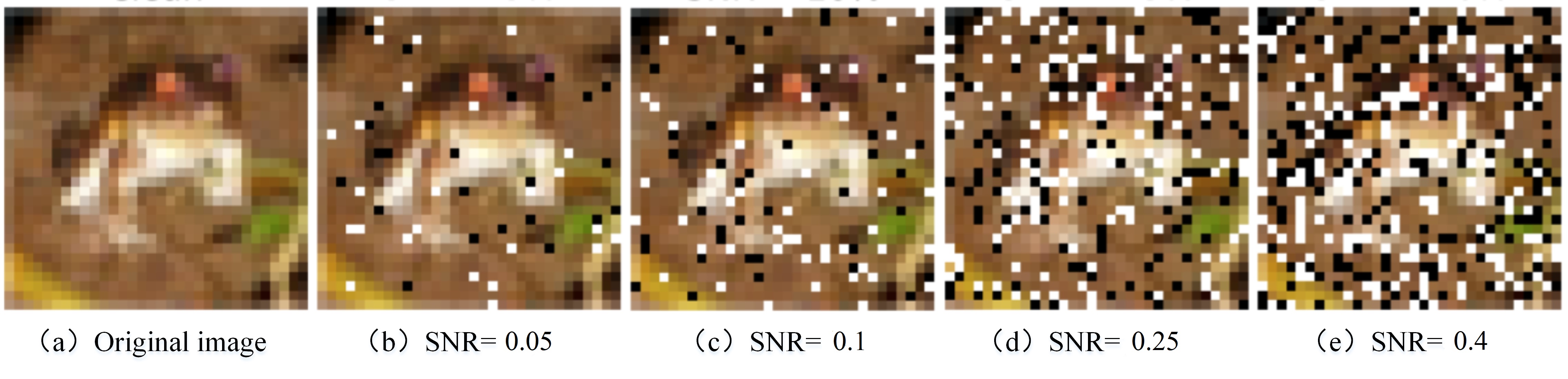}
\caption{Noisy images with different SNR levels.}\vspace{-0.15in}\label{fig7}%
\end{figure}

In this experiment, salt-and-pepper noises are leveraged~\cite{b40}. The noise is simulated by adding white (salt) or black (pepper) noises into the original RGB images with a parameter of signal-to-noise ratio (SNR). In our experiment, the SNR is set as 0.4 for each image. Fig. \ref{fig7} shows an example for nosiy images with different SNR levels. The noise rate on the whole data is set as $20\%$ and $40\%$, respectively.

\begin{figure}
\centering
\includegraphics[width=0.75\linewidth]{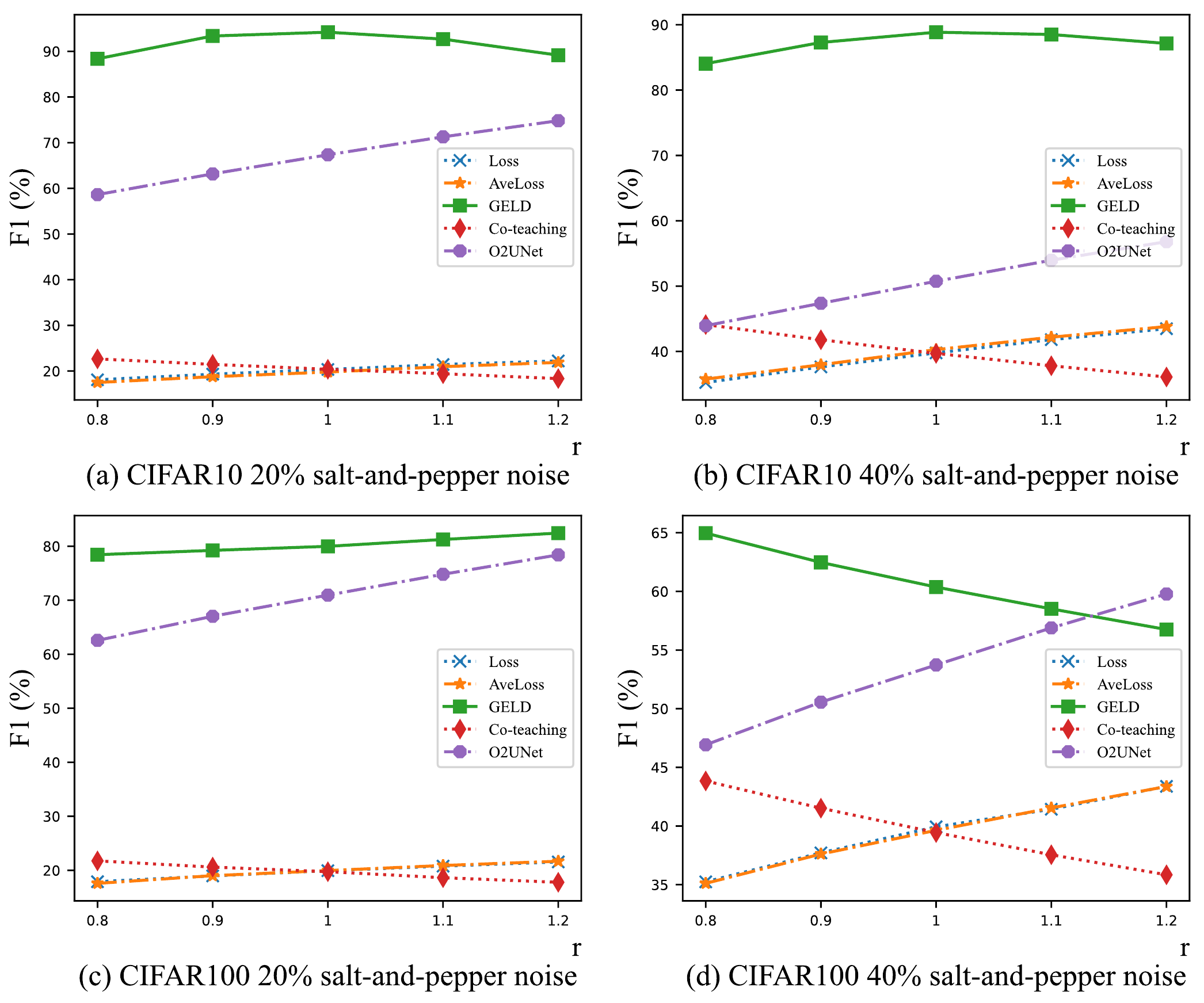}
\caption{The F1 scores ($\%$) of the competing methods on CIFAR10 and CIFAR100 under different rates of salt-and-pepper noise.}
\label{fig8}\vspace{-0.1in}
\end{figure}

Fig. \ref{fig8} shows the performances of the competing methods on CIFAR10 and CIFAR100, respectively. The settings of competing methods remain unchanged compared with the label noise experiments. Our method GELD still achieves the highest F1 values under all the noise rates. Few noisy data were detected by methods that rely merely on the loss.

\subsection{Measurement under Small-margin Data Detection}
\indent

A sample with a small margin (the distance to the oracle decision boundary) is considered to be hard in learning \cite{b3}. This experiment evaluates a learning difficulty measure in terms of the detection of small-margin samples.

In this experiment, four UCI~\cite{b39} data sets are used, namely, Iris, Mammographic, Haberman, and Abalone. To better construct the ground-truth, only binary classification is considered. Only two categories are selected for both Iris (the ``Setosa" and ``Versicolour" categories) and Abalone (the ``9" and ``10" categories). Mammographic and Haberman contain only two categories. The details of used data in this experiment are presented in Table \ref{tab2}. The classical margin-based learning method SVM~\cite{b32} is used to construct the ground truth, i.e., the small-margin samples. Specifically, the SVM with RBF kernel is used. Two parameters $C$ and $g$ are searched in \{$10^{-3}, 10^{-2}, 10^{-1},1,10,10^{2},10^{3}$\} and \{$10^{-3}, 10^{-2}, 10^{-1},1,10,10^{2},10^{3}$\}, respectively, via five-fold cross-validation. The optimal parameter setting is used and the SVM is trained on the whole training set. Constantly, the margin of each sample is calculated as the ground-truth difficulty. The margin is $yf(x)$ for the sample $x$, where $y$ is the label and $f(x)$ is the output of the kernel SVM. Let $N$ be the \#Instances. The top-$N*v$ samples with small margins are selected as the ground-truth samples to detect.

As the base network ResNet-34 is inappropriate in this experiment, a three-layer perception with the Sigmoid activation function is used as the base network. The number of epoch is set as 10000. Its hyper-parameters are also pursed via five-fold cross-validation. Considering that MentorNet and Co-teaching are quite complex for this scenario, they are not compared in this experiment. The competing methods include Loss, AveLoss, O2UNet, and our GLED. The evaluation criteria and the whole calculate scheme follow the setting in the previous experiments. The value of $v$ is set as 20\%, 40\%, and 60\%, respectively; $r$ is set as one.

\begin{table}
\begin{center}
\begin{minipage}{234pt}
\caption{Details of the four UCI data sets.}\label{tab2}%
  \begin{tabular}{cccc}
    \toprule
    Data set & \#Dimensions & \#Classes & \#Instances\\
    \midrule
    Iris &   4 & 2 & 100\\
    Mammographic &   5 & 2 & 961\\
    Haberman &   3 & 2 & 360 \\
    Abalone &   8 & 2 & 1323\\
  \bottomrule
\end{tabular}
\end{minipage}
\end{center}
\end{table}

\begin{figure}
\centering
\includegraphics[scale=0.4]{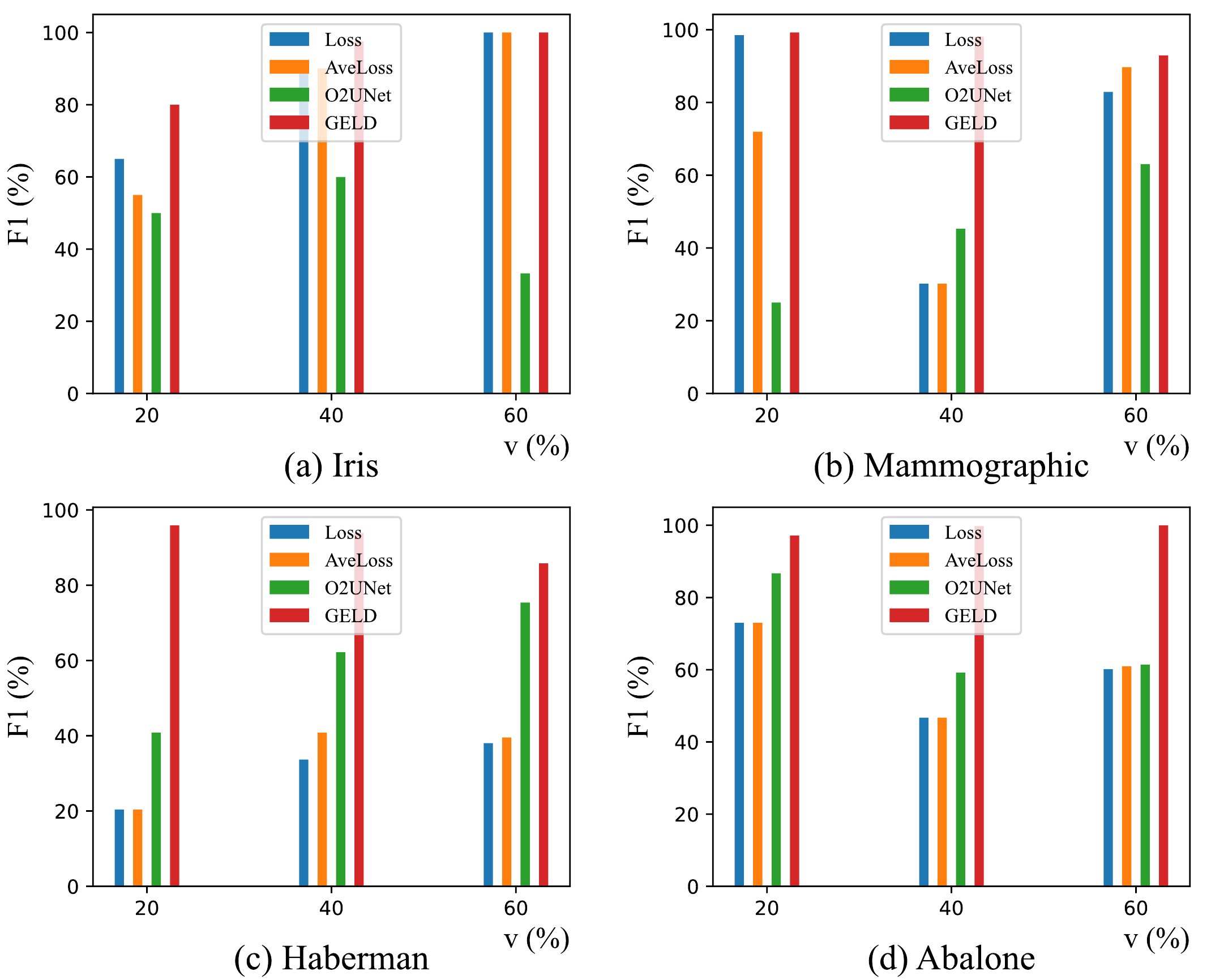}
\caption{Histograms of F1 scores (\%) under various data sets using margin as ground-truth.}\vspace{-0.1in}
\label{fig9}
\end{figure}

The results are shown in Fig. \ref{fig9}. Our approach GELD achieves the highest F1 values on all data sets. In addition, GELD is stable across different $v$s and different data sets. By contrast, the other methods are not stable. 

\subsection{Measurement under Epistemic Uncertainty Detection}
\indent

In this experiment, a data corpus containing epistemic uncertainty is required. An image aesthetic assessment data corpus, namely, AVA benchmark image aesthetic data corpus~\cite{b42}, is then utilized and each photo receives multiple rating scores from multiple different users. The variance of user rating scores of a photo reflects the epistemic uncertainty on the photo using each user as a gold model. A large variance indicates a large uncertainty for a photo. Specifically, the former 50,000 samples in the AVA corpus are downloaded and each sample receives 210 rates on average counting from 1 to 10. We use ResNet-101 as the base network as the image quality of AVA is much higher than CIFAR10 and CIFAR100. Given that the Bayesian Neural Network (BNN)~\cite{b20} is particularly designed for modeling uncertainty, we import BNN as one of the competing methods, and its outputting confidence coefficient is used to detect uncertain photos. Therefore, the competing methods include Loss, AveLoss, O2UNet, BNN, and our GELD.

The top-$50,000*v$ photos with high uncertain scores are taken as the objective samples to detect. The $v$ values are set as 20\%, 40\%, and 60\%. The $r$ value is set as one. In the BNN method, the dropout strategy described in~\cite{b20} is followed. The parameter setting of ResNet-101 reported in~\cite{b38} is adopted. The photos are resized into shape of (3, 192, 192) to fit in the ResNet-101.

\begin{table}
\begin{center}
\begin{minipage}{234pt}
\caption{$F1$ scores (\%) using the variance of scores as standard}\label{tab3}%
  \begin{tabular}{ccccccc}
    \toprule
    Method & Loss & AveLoss & O2UNet & BNN & GELD\\
    \midrule
    $20\%$ &   14.64 & 54.05& 33.27 & 62.71&\textbf{63.96}\\
    $40\%$ &   28.38 & 33.33& 35.67 & 54.30&\textbf{78.39}\\
   $60\%$ &   30.23 & 45.79& 38.20 & 78.21&\textbf{79.11}  \\
  \bottomrule
\end{tabular}
\end{minipage}
\end{center}
\end{table}

\begin{figure}[htbp]
\centering
\includegraphics[scale=0.14]{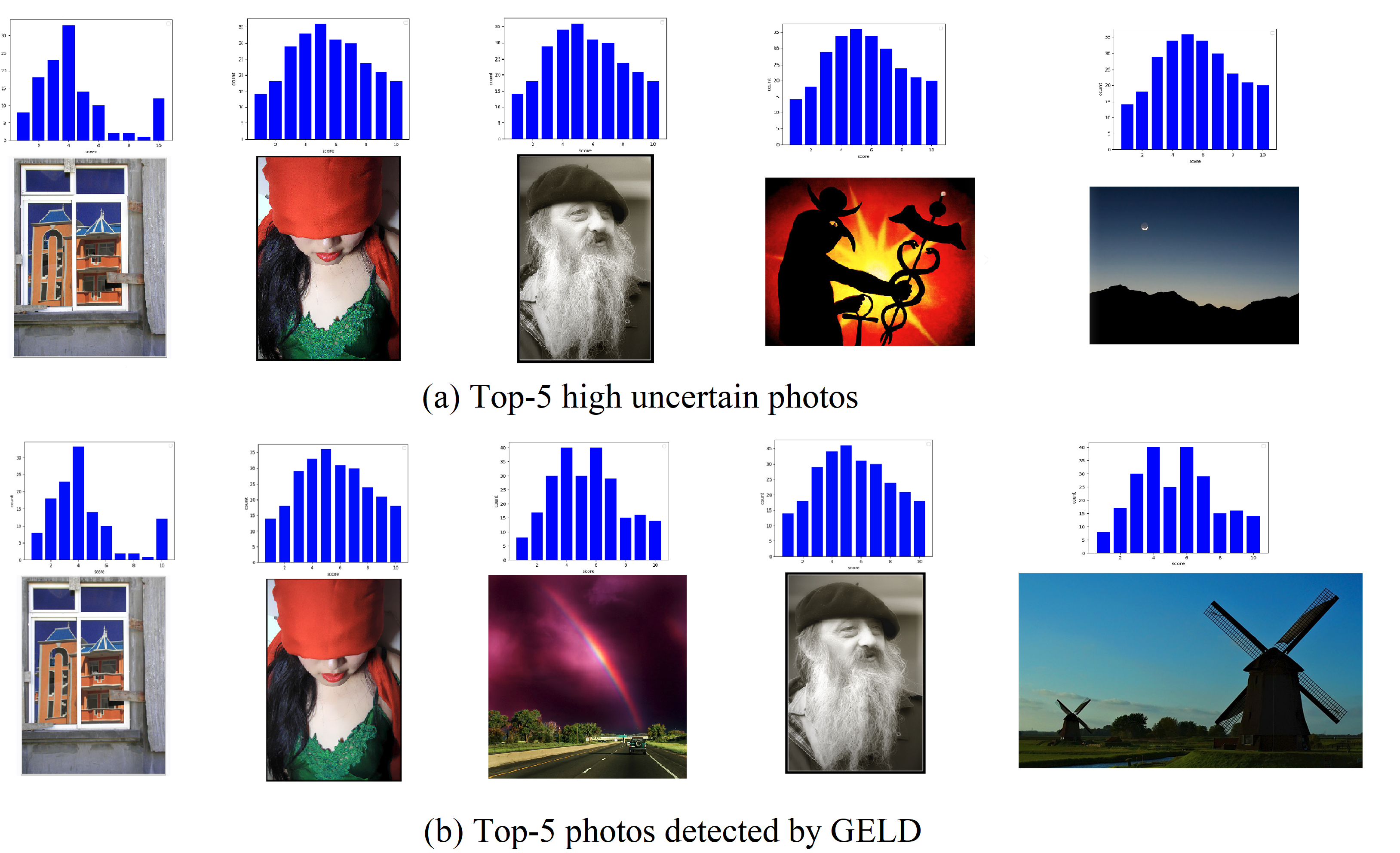}
\caption{True highest uncertain photos and our detection results. The histograms of user ratings are also presented.}\vspace{-0.1in}
\label{fig10}
\end{figure}

\begin{figure}[htbp]
\centering
\includegraphics[scale=0.1]{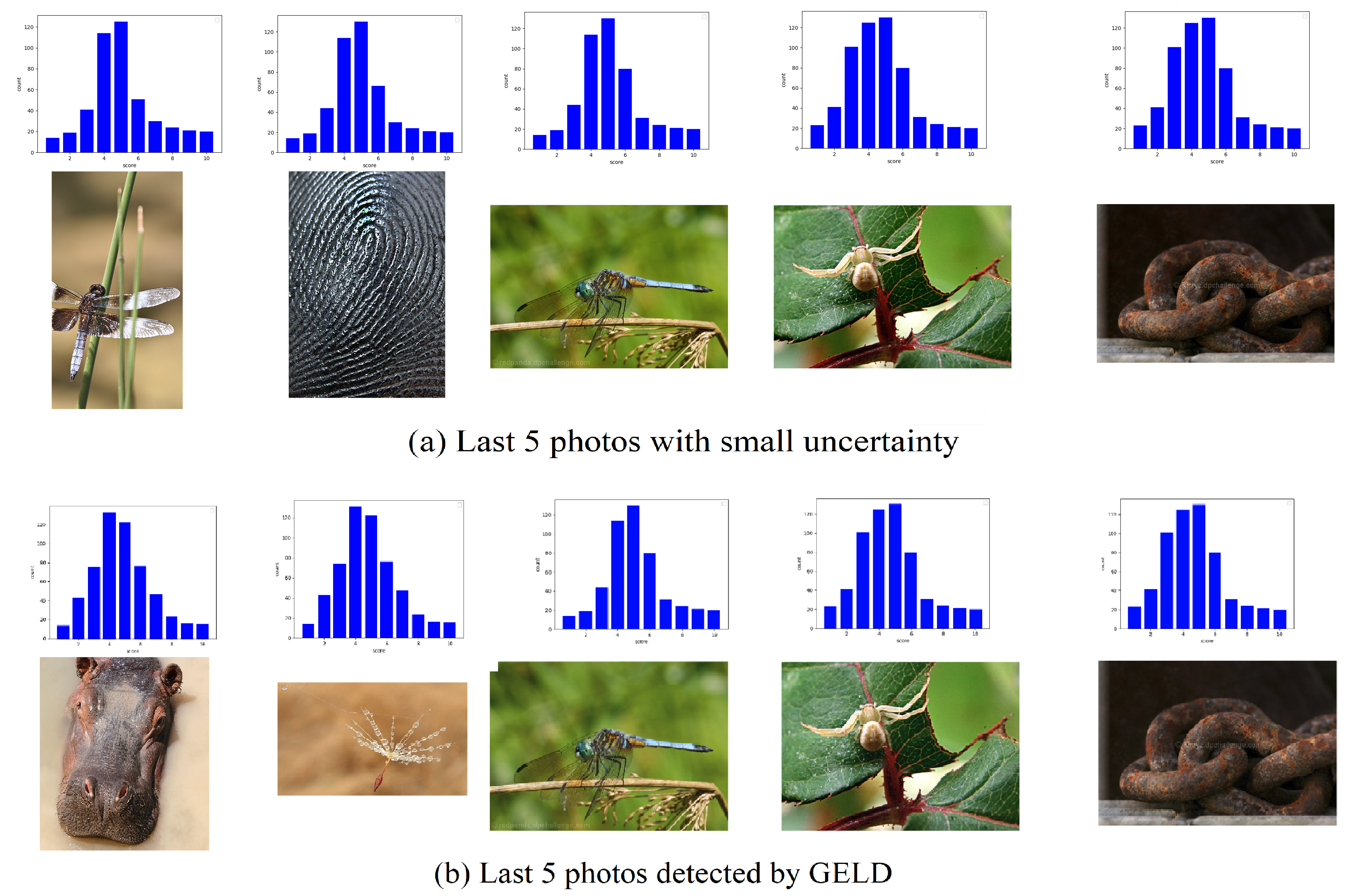}
\caption{True lowest uncertain photos and our detection results. The histograms of user ratings are also presented.}\vspace{-0.1in}
\label{fig11}
\end{figure}

Table \ref{tab3} shows the F1 scores of the competing methods in high uncertain sample detection. GELD still performs the best and slightly outperforms BNN. The rest of the three methods poorly perform in this detection task. 
Fig.~\ref{fig10} shows the top-5 high uncertain photos and the top-5 photos detected by our GELD approach. The aforementioned figure (Fig.~\ref{fig11}) also shows the last five photos with small uncertainty and the last five photos detected by our GELD.

\subsection{Measurement for Imbalance Data}
\indent

The long-tail versions of the CIFAR10 and CIFAR100 are used in this experiment. Buda et al.~\cite{b34} compiled a series of data sets under different imbalance ratios. The two data sets under the $20:1$ ratio for CIFAR10 and CIFAR100  are used. There is no ground-truth information for the learning difficulties because the head categories can also contain difficult samples. Consequently, we only plot the histograms of the numbers of hard samples detected by the competing methods in the head and tail categories. The competing methods are Loss, AvgLoss, O2UNet, VOG, and our proposed GLED. The base network and the concerning setting in part \textit{A} are followed.
\indent

\begin{figure}[htbp]
\begin{center}
\begin{minipage}{234pt}
\includegraphics[scale=0.5]{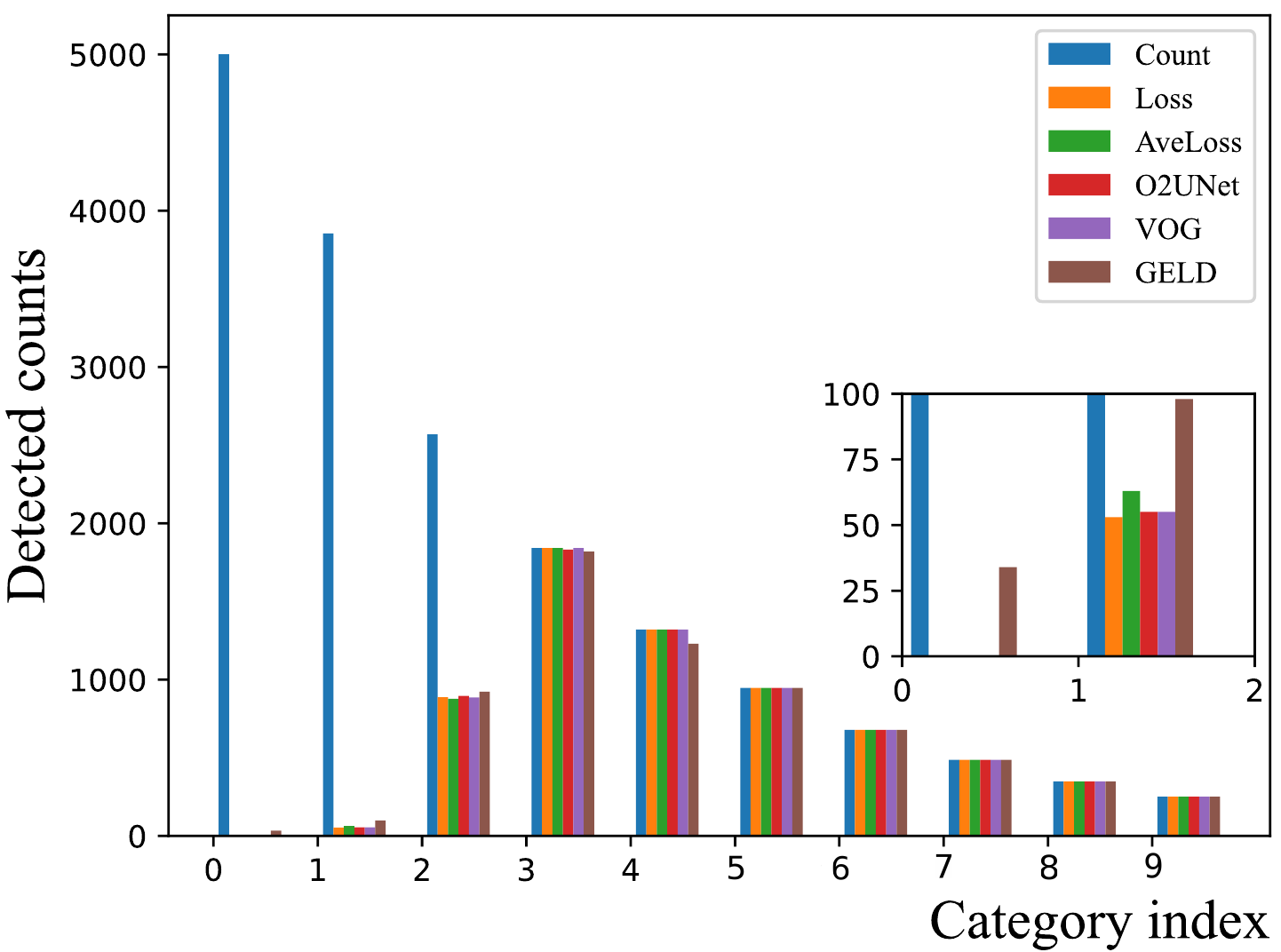}
\caption{ Histogram of CIFAR10-LT top-$40\%$ detected hard samples.}\label{fig12}%
\end{minipage}
\end{center}
\end{figure}

\begin{figure}[htbp]
\begin{center}
\begin{minipage}{234pt}
\includegraphics[scale=0.5]{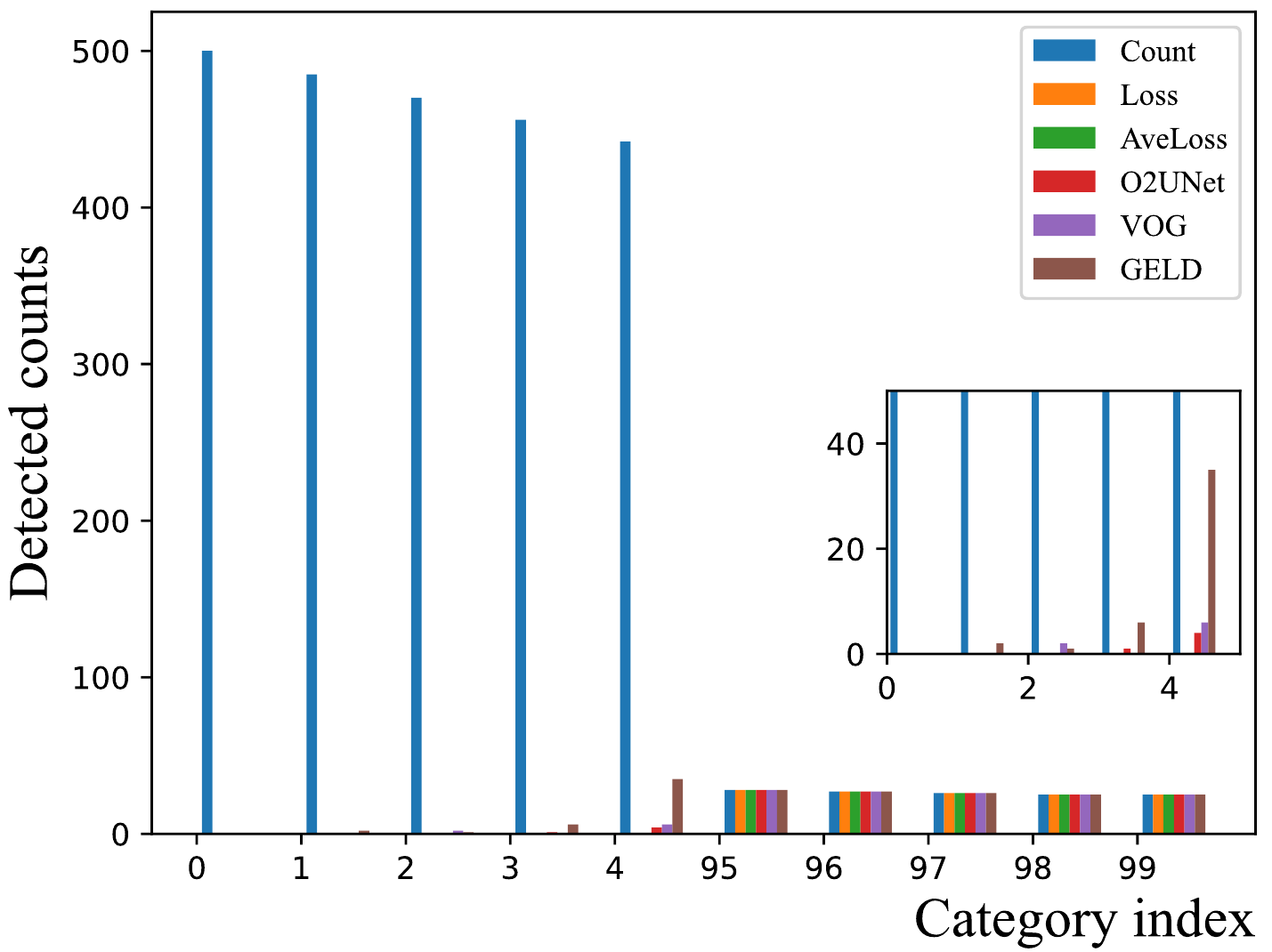}
\caption{ Histogram of CIFAR100-LT top-$40\%$ detected hard samples.}\label{fig13}%
\end{minipage}
\end{center}
\end{figure}

The top-40\% samples detected by each competing methods are regarded as their detected hard samples. Figs. \ref{fig12} and \ref{fig13} show the histograms of the detected hard samples by each method on CIFAR10 and CIFAR100 (the top five head categories and the last five tail categories), respectively. All the competing methods identically behave on the tail categories on both data corpora. This condition is reasonable and accords well with the primary motivation for imbalance learning that samples in tail categories are hard to learn. There are slight differences between our GELD and other competing methods on the head categories. The numbers of hard samples detected by GELD are larger than those of other methods, which is reasonable because head categories still contain hard samples.
\begin{figure}[htbp]
\centering
\includegraphics[scale=0.3]{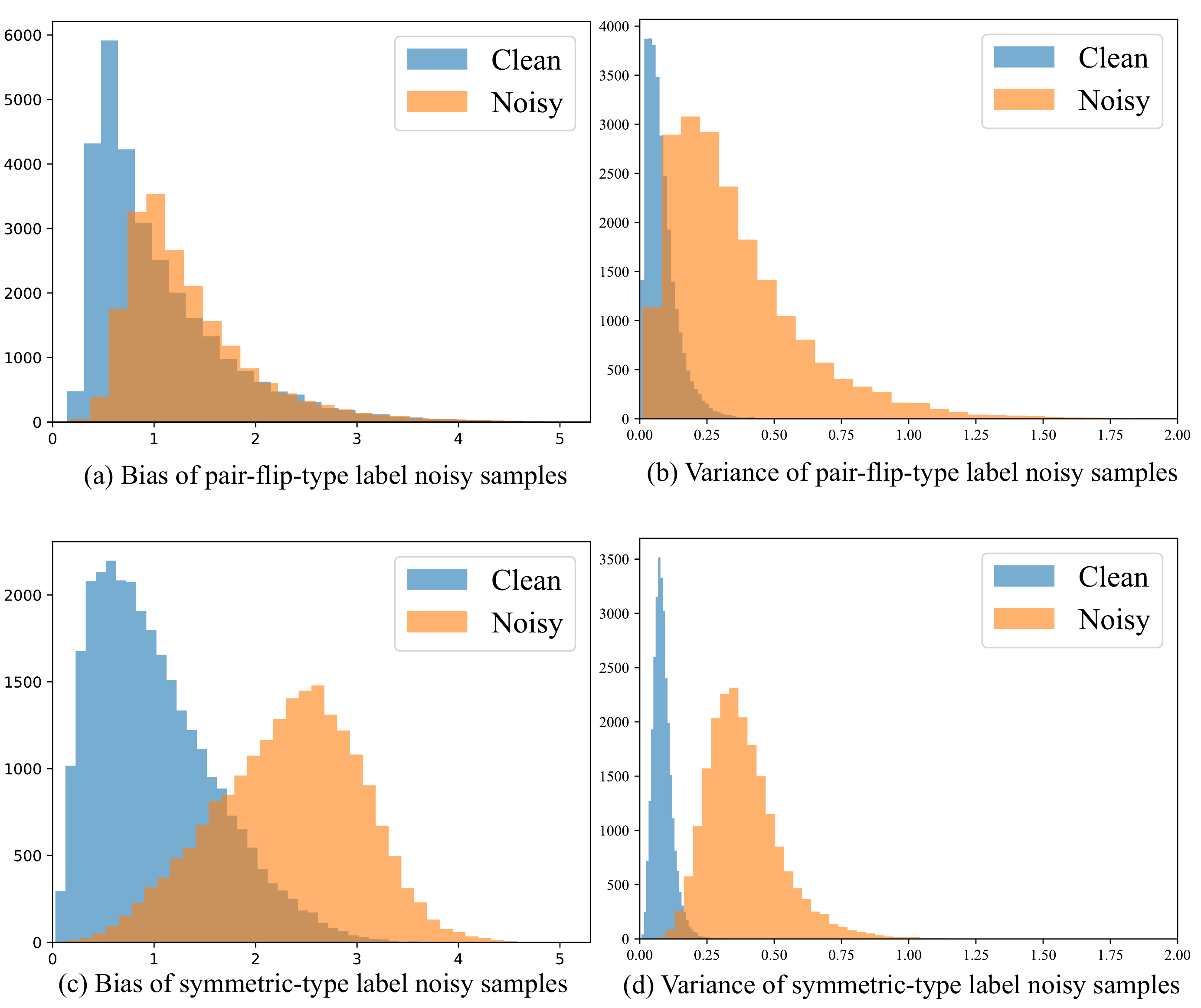}
\caption{Histograms of bias and variance values calculated by GELD under different types of label noise ($v = 40\%$).}
\label{fig14}\vspace{-0.2in}
\end{figure}

\subsection{Discussion}
\indent

The above experiments on the four scenarios, namely, noise detection, small-margin sample detection, uncertain sample detection, and hard sample detection in imbalance learning, verify the superiority of the proposed GELD approach over existing classical and state-of-the-art methods. As previously introduced, the primary difference between GELD and many existing methods lies in that GELD explicitly considers variances. Fig. \ref{fig14} shows the histograms of the bias values and the variance values achieved by GELD on clean samples and label noisy samples under both pair-flip and symmetric noise types on CIFAR10. The variance values between clean and noisy samples are also considerably distinct as shown in Figs. \ref{fig14} (b) and (d), which demonstrates the usefulness of the variance term utilized in our GELD method. In addition, the difference consist in histograms of noisy and clean samples under the pair-flip noises is trivial, which rationalises the poor performances of the loss-based methods such as O2UNet.



Although good results are achieved, GELD is only an appropriation for theoretical difficulty in Definition 1. We evaluate the robustness of the method in terms of the variations on the two key parameters, namely, $K$ and $M$. Table \ref{tab4} shows the performances of pair-flip noise detection ($r=1$) on CIFAR10 and CIFAR100. The results show that the performances of GELD are stable when the value of $(K,M)$ is set in \{(4, 5), (5, 6), (5, 8), (5, 10)\}.

\renewcommand{\arraystretch}{1.1}
\begin{table}
\begin{center}
\begin{minipage}{234pt}
\caption{$F1$ scores (\%) with various values of $K*M$.}
  \label{tab4}
  \begin{tabular}{cccccc}
    \toprule
   &$K*M$ &20 (4$\times 5$)& 30 ($5\times 6$) &40 ($5\times 8$) &  50 ($5\times 10$) \\
    \midrule
     \multirow{3}*{\rotatebox{90}{CIFAR10}} & $20\%$ & 96.83&  97.51 & 96.91 & 97.04\\
    \cline{2-6} & $40\%$ &94.12 &  95.77& 96.10&  96.34\\
     \cline{2-6} & $60\%$ & 94.81 &  95.82& 94.95 & 95.34 \\
  \midrule
    \midrule
     \multirow{3}*{\rotatebox{90}{CIFAR100}}  & $20\%$ & 90.36 & 91.00 & 92.67&  91.46\\
     \cline{2-6} & $40\%$ & 91.78& 92.48 & 91.86& 91.04 \\
     \cline{2-6} & $60\%$ & 90.02& 92.00& 91.17&  90.89\\
  \bottomrule
\end{tabular}
\end{minipage}
\end{center}
\end{table}

 We are interested in whether a simple model can also obtain better results. A simple network, namely, AlexNet~\cite{b35}, is used as the basic leaner in pair-flip noisy label. The base network and setting of other methods follow the previous setting of the corresponding experiments in part \textit{A}. The training setting of AlexNet in~\cite{b43} is followed, and the output-size is modified into 100 while training CIAFR100. The results are shown in Table \ref{tab5}. GELD (AlexNet) is inferior to GELD (ResNet-34). However, it is comparable to Co-teaching and outperforms the rest of the methods.

\begin{table}
\begin{center}
\begin{minipage}{234pt}
\caption{$F1$ scores (\%) of the competing methods plus GELD using AlexNet as base model.}
  \label{tab5}
  \begin{tabular}{cccccccc}
    \toprule
    Method &\multicolumn{3}{c}{CIFAR10} & \multicolumn{3}{c}{CIFAR100}\\
    \midrule
    Label noise rate & $20\%$ & $40\%$& $60\%$& $20\%$& $40\%$& $60\%$\\
    \midrule
    GELD (ResNet-34) &{97.51} &{95.77}&{95.82}&{91.00}&{92.48}&{92.00}\\
    O2UNet & 92.31&80.76&33.69&61.29&54.03&46.98\\
    Co-teaching& 92.63&89.45&58.02&86.22&89.77&88.90\\
    MentorNet& 88.02&67.96&47.14&73.00&51.49&60.09\\
    AveLoss & 81.39&79.23&51.98&77.38&74.18&79.17\\
    Loss& 76.69&76.12&50.53&75.29&74.15&79.75\\
    GELD (AlexNet)& 82.58&88.77&83.54&82.96&86.51&84.39\\
  \bottomrule
\end{tabular}
\end{minipage}
\end{center}
\end{table}

\begin{table}
\begin{center}
\begin{minipage}{190pt}
\caption{Time cost of different complex methods.}
  \label{tab6}
  \begin{tabular}{cc}
    \toprule
    Method & Time cost (hours)\\
    \midrule
    O2UNet & 7.95\\
    GELD (1 GPU) & 11.11\\
    GELD (2 GPUs) & 5.67\\
    GELD (3 GPUs) & 3.71\\
    GELD (4 GPUs) & 2.83 \\
  \bottomrule
\end{tabular}
\end{minipage}
\end{center}
\end{table}

A large real-world data set, namely, Clothing1M~\cite{b33}, is used to further evaluate the performance of our GELD measure in terms of image classification. There are 14 classes in Clothing1M containing 1,000,000 training images with real noisy labels, 48,000 training samples verified to be clean, and 10,000 testing images. The 48,000 clean training samples are used as the validation data for Clothing1M. Clothing1M has a noise proportion
of $38\%$ approximately. ResNet-101 is used and the settings of the network in~\cite{b17} are employed. $(K,M)$ are set as $(50,50)$. Each model is learned for 50 epochs for GELD. The model is selected using the 48,000 clean training samples. The learning rate remains constant as $1e-6$. The batch size is set as 16 during the GELD calculation. Other hyper-parameters follow the settings in~\cite{b17}. The top $10\%$ samples with the highest $Err(x_i,\lambda_h)$ values are removed as detected noisy samples. Remaining samples are used to learn the final image classifier. The batch size is set as 128 and the maximum epoch is set as 10 during this procedure. The settings of other methods in ~\cite{b17} are followed. Table \ref{tab7} shows the comparison of classification accuracies ($\%$) among the four competing methods. GELD outperforms the other three methods. Results of methods beside GELD are directly from the O2UNet study~\cite{b17}.

\begin{table}
\begin{center}
\begin{minipage}{200pt}
\caption{Classification accuracy (\%) on Clothing1M}\label{tab7}%
  \begin{tabular}{cccc}
    \toprule
    MentorNet & Co-teaching & O2UNet & GELD\\
    \midrule
    79.30 & 78.52 & 82.38 &\textbf{82.94}\\
  \bottomrule
\end{tabular}
\end{minipage}
\end{center}
\end{table}
The computational cost of GELD is relatively high as $K*M$ models should be trained. However, it is still smaller than another SOTA method O2U-Net. Moreover, several ways can significantly reduce the complexity. First, the task can be performed in parallel. Four NVIDIA GeForce RTX 3090 GPUs are used in our experiments. The average time costs for our GELD using different number of GPUs on CIFAR10 using ResNet-34 as base network are shown in Table \ref{tab6}. The settings follow the pair-flip label noise experiments' in part \textit{A} with $(K, M)$ is set as (5, 6). The time consumption is considerably reduced when GELD is run on more GPUs in parallel. The time cost of O2UNet is also large. More over, O2UNet cannot be performed in parallel. Second, a relatively small training set instead of the entire can be used when dealing with large corpora. Third, a dropout-based strategy (like the quantifying uncertainties in BNN) can also reduce the time cost. To sum up, the time complexity of GELD doesn't hinder its applications based on these strategies.

Although four key factors, namely, data quality, sample margin, uncertainty, and category distribution, are summarized and the proposed method achieves quite competing performance, a directly theoretical connection between the four factors and the learning difficulty is not established in this study. We leave this theoretical investigation as our future work.
\section{Conclusion}
\indent

This study has conducted a comprehensive investigation on learning difficulty of data in machine learning. We established a theoretical definition of learning difficulties of data based on the bias-variance trade-off on generalization error. The well discussed and explored concepts, easy and hard samples, are formally described based on the theoretical definition and the associated difficulty coefficients. Influential factors of learning difficulty are summarized and correlations between generalization error, model complexity, and influential factors are surveyed and analysed. A practical measure, namely, the generalization  error-based learning  difficulty  (GELD)  measurement, is then proposed in virtue of influential factors to calculate the learning difficulty of each training sample. Finally, the properties of the weighted learning strategy are presented and three classical methods are explained on the basis of the theoretical formalization. Extensive experiments validate the effectiveness of our proposed measure, which outperforms existing state-of-the-art methods under different scenarios considering concluded influential factors.

This study conducts an attempt to establish a theory for learning difficulty of samples. Our future work aims to reveal the mathematical correlations between the theoretical definition (i.e., optimal model complexity) and the measure (i.e., generalization error) for learning difficulty.

\backmatter

\section*{Declarations}
\subsection*{Funding}
Not applicable.
\subsection*{Conflict of interest}
We declare that we have no financial and personal relationships with other people or organizations that can inappropriately influence our work, there is no professional or other personal interest of any nature or kind in any product, service and/or company that could be construed as influencing the position presented in, or the review of, the manuscript entitled.
\subsection*{Ethics approval}
Not applicable.
\subsection*{Consent to participate}
Not applicable.
\subsection*{Consent for publication}
Not applicable.
\subsection*{Availability of data and materials}
The data sets used or analysed during the current study are available from the corresponding author on reasonable request. All data generated or analysed during this study are included in this published article.
\subsection*{Code availability}
All codes are uploaded to Github source repository:\\
\url{https://github.com/Weiyao619/GELD.git}
\subsection*{Authors' contributions}
Wu Ou contributed to the conception of the study;\\
Wu Ou and Zhu Weiyao contributed significantly to analysis and manuscript preparation;\\
Zhu Weiyao performed the experiment;\\
Su Fengguang performed the data analyses and wrote the manuscript;\\
Deng Yingjun helped perform the analysis with constructive discussions.

\bigskip

\begin{appendices}
\setcounter{equation}{0}
\renewcommand\theequation{A.\arabic{equation}}
\setcounter{figure}{0}
\renewcommand\thefigure{A.\arabic{figure}} 
\section{Calculation of Model Complexity}
\indent

The model complexity used in this study is on the basis of minimum description length (MDL)~\cite{b60} and the Kolmogorov complexity~\cite{b61}. MDL and Kolmogorov complexity are combined and used to describe the model complexity under various learning tasks in~\cite{b66}.

Let $\textbf{w} =  (w_1,..,w_{d-1})^T$ be the model parameter of a regression model where $d$ signifies the dimension of inputs. Let $p(\textbf{w})$ be the probability density of \textbf{w}.
According to MDL, the model complexity expectation $c(\textbf{w})$ is the expectation of model complexity
$$m(\textbf{w}) = -\log p(\textbf{w}),$$ over different training set $T$, i.e., $c(\textbf{w}) = \mathbb{E}_T[m(\textbf{w})]$. Suppose that each component of $\textbf{w}$ is independent to each other and follows the identical Gaussian distribution $w_i \sim \mathscr{N}(0,\sigma^2).$ Therefore, the model complexity defined on the basis of MDL equals to 
\begin{equation}
\label{A.1}
    \begin{aligned}
      m(\textbf{w}) & = -\log p(\textbf{w})\\
      & = - \sum_{i =1}^{d} \log p(w_i) \\
      & = - \log \prod_{i =1}^{d} p(w_i) \\
      & = - \log \prod_{i =1}^{d} [\dfrac{1}{\sqrt{2\pi}\sigma}\exp(-\dfrac{w_i^2}{2\sigma^2})]\\
      & = \sum_{i =1}^{d} \dfrac{w_i^2}{2\sigma^2} + d\log(\sqrt{2\pi}\sigma)\\
      & = \dfrac{\Vert\textbf{w}\Vert_2^2}{2\sigma^2}+ d\log(\sqrt{2\pi}\sigma)
    \end{aligned}
\end{equation}
When the construction of the basic learner and $\sigma^2$ is fixed, the model complexity only concerns $\Vert\textbf{w}\Vert_2^2$.

\indent

Under the ridge regression, with the input denoted as $x$ and its label denoted as $y$, the objective function is $$\mathscr{L} = \sum_{n=1}^{N} l(f(x;\textbf{w}),y) + \lambda \Vert \textbf{w}\Vert_2^2,$$ and the estimated parameters are given by $$\hat{\textbf{w}}(\lambda)=(x^Tx+\lambda I)^{-1} x^Ty.$$ 

Based on the Assumption \ref{assump1}, the variance term is an increasing function with respect to the model complexity, and as the loss function normally used during the training procedure is in fact the bias term in the bias-variance trade-off, the generalization error can be estimated by the following form:
\begin{equation*}
    \begin{aligned}
      \mathscr{L} &= \sum_{n=1}^{N} l(f(x;\textbf{w}),y) + m(\textbf{w})\\
      & = \sum_{n=1}^{N} l(f(x;\textbf{w}),y) + \dfrac{\Vert\textbf{w}\Vert_2^2}{2\sigma^2}+ d\log(\sqrt{2\pi}\sigma)\\
      & \sim \sum_{n=1}^{N} l(f(x;\textbf{w}),y) + \lambda \Vert\textbf{w}\Vert_2^2
    \end{aligned}
\end{equation*}

Accordingly, an enlargement of $\lambda$ leads to a reducing of $\Vert\textbf{w}\Vert_2^2$, and moreover, a lower model complexity. The above analysis indicates that the MDL-based model complexity well explains the ridge regression. However, the former calculation is based on the identical distribution assumption for each $w_i$. When the model is a polynomial function, the contributions of each component of $\textbf{w}$ to the whole model complexity are not identical. For example, when using a polynomial function $g(x) \sim \mathscr{O}(3)$ to perform ridge regression, $w_3$ should contributes more to the model complexity comparing to $w_0$. Therefore, an identical distribution for all components of $\textbf{w}$ is unreasonable. A more reasonable assumption is that $w_i \sim \mathscr{N}(0,\sigma_i^2)$ with the condition that $\sigma_i^2 < \sigma_j^2 $ if $i>j$. From Eq.~(\ref{A.1}), we have 
\begin{equation}
\label{A.2}
    \begin{aligned}
      m(\textbf{w}) & = -\log p(\textbf{w})\\
      & = - \sum_{i =1}^{d} \log p(w_i) \\
      & = - \log \prod_{i =1}^{d} p(w_i) \\
      & = - \log \prod_{i =1}^{d} [\dfrac{1}{\sqrt{2\pi}\sigma_i}\exp(-\dfrac{w_i^2}{2\sigma_i^2})]\\
      & = \sum_{i =1}^{d} \dfrac{w_i^2}{2\sigma_i^2} + d\log(\sqrt{2\pi}\sigma_i)
    \end{aligned}
\end{equation}

In our practical calculation, let $\sigma_i^2 = (\dfrac{d}{i}\sigma)^2$. Denoting $\hat{\textbf{w}}_t(\lambda)$ as parameters of model learnt on training set $T_t$. Ignoring the constant term, the model complexity becomes
\begin{equation*}
    m(\hat{\textbf{w}}_t(\lambda)) =   \sum_{i =1}^{d} (\dfrac{i}{d} \hat{w}_{t,i}(\lambda))^2.
\end{equation*}
Given $M$ training sets, the model complexity expectation is
\begin{equation}
\label{A.4}
    c(\hat{\textbf{w}}(\lambda)) =  \dfrac{1}{M}\sum_{i =1}^{M} \sum_{i =1}^{d} (\dfrac{i}{d} \hat{w}_{t,i}(\lambda))^2.
\end{equation}

\section{Proofs of Propositions}

\subsection{Proof of Proposition1}

\begin{proof}
According to Assumption 1, the partial derivatives of the generalization error with respect to $c$ is negative before the generalization error achieve the minimum. Accordingly, if 
\begin{equation*}
\footnotesize
    \dfrac{\partial \overline{Err}(\lambda_{h})}{\partial c}\vert_{c=c^*} < 0,
\end{equation*}
then $c'^*$ should be larger than $c^*$.
\end{proof}

\subsection{Proof of Proposition3}

\begin{proof}
\begin{small}
\begin{equation*}
\begin{aligned}
  Err^{\text{w}}(\lambda_{h}) & = \sum_{y \in \Omega_Y}P(y) \int_{x\in \Omega_X} \omega\  Err(x,\lambda_{h})p(x\vert y)\mathrm{d}x \\
    & = \sum_{y \in \Omega_Y} P(y) \int_{x \in \Omega_X^r} \omega\   Err(x, \lambda_{h})p(x\vert y) \mathrm{d}x \\
    & + \sum_{y \in \Omega_Y} P(y) \int_{x \in \Omega_X / \Omega_X^r} Err(x, \lambda_{h})p(x\vert y)\mathrm{d}x \\
    & = Err(\lambda_h) + \sum_{y \in \Omega_Y} P(y) \int_{x \in \Omega_X^r} (\omega - 1) Err(x, \lambda_{h})p(x\vert y)\mathrm{d}x
\end{aligned}.
\end{equation*}
\end{small}
Note that $\footnotesize{ \dfrac{\partial Err(\lambda_{h})}{\partial c} \vert_{c = c^*} = 0}$.
Given that $\mathcal{LDC}(x) > 1$,  $\footnotesize \dfrac{\partial Err(x, \lambda_{h})}{\partial c} \vert_{c = c^*} < 0,\ \forall x \in \Omega^r.$
With $\omega > 1$, we have $$\footnotesize \dfrac{\partial Err^{\text{w}}(\lambda_{h})}{\partial c}\vert_{c = c^*} < 0.$$
According to Proposition 1, the new optimal model complexity $c'^*$ will be larger than $c^*$.
\end{proof}

\subsection{Proof of Corollary1}

\begin{proof}
The optimal complexity $c^*$ under the original weights $\omega$ can be theoretically inferred under the (original) weighted distribution $P_1 \sim  \omega P$. The learning with new weights $\tilde{\omega}$ equals to the learning with the weights $\tilde{\omega} / \omega$ for each sample in $\Omega^r$ under the distribution $P_1$. Because the new weights are larger than the original weights on $\Omega^r$, $\tilde{\omega} / \omega$ is larger than one on $\Omega^r$. According to Proposition 3, the new optimal complexity becomes larger.
\end{proof}

\subsection{Proof of Proposition5}
\begin{proof}
Let $\Omega^e$ and $\Omega^h$ be the regions containing the easy and hard samples according to $c^*$, respectively.   
 \begin{footnotesize}
\begin{equation*}
 \begin{aligned}
   \dfrac{\partial Err^{\text{w}_0}(\lambda_{h})}{\partial c}\vert_{c=c^*} = & \sum_{y \in \Omega_Y} P(y) \int_{x \in \Omega_X^e} \omega_0(x)\dfrac{\partial Err(x,\lambda_{h})}{\partial c}\vert_{c=c^*} p(x\vert y) \mathrm{d}x \\
   & + \sum_{y \in \Omega_Y} P(y) \int_{x \in \Omega_X^h} \omega_0(x)\dfrac{\partial Err(x,\lambda_{h})}{\partial c}\vert_{c=c^*} p(x\vert y) \mathrm{d}x \\
   =&0
 \end{aligned}.
 \end{equation*} 
\end{footnotesize}
 Let $\omega^*_e = \max \limits_{x\in \Omega^e} \omega(x)$ and $\omega^*_h = \min \limits_{x\in \Omega^h} \omega(x)$. Moreover, $\omega^*_e \leq \omega^*_h$. We have
 \begin{footnotesize}
  \begin{equation*}
     \begin{aligned}
       \dfrac{\partial Err^{\text{w}}(\lambda_{h}) }{\partial c} =  & \sum_{y \in \Omega_Y} P(y) \int_{x \in \Omega_X^e} \omega(x) \dfrac{\partial  Err^{\text{w}_0}(x,\lambda_{h})}{\partial c} p(x\vert y) \mathrm{d}x \\
       & + \sum_{y \in \Omega_Y} P(y) \int_{x \in \Omega_X^h} \omega(x) \dfrac{\partial Err^{\text{w}_0}(x,\lambda_{h})}{\partial c} p(x\vert y) \mathrm{d}x
     \end{aligned}
 \end{equation*}
 \end{footnotesize}
 Note that 
 \begin{equation*}
 \footnotesize
     \begin{aligned}
       \dfrac{\partial Err^{\text{w}_0}(x, \lambda_{h})}{\partial c}\vert_{c=c^*} &> 0, \forall \ x \in \Omega_X^e\\
       \dfrac{\partial Err^{\text{w}_0}(x, \lambda_{h})}{\partial c}\vert_{c=c^*} &< 0, \forall \ x \in \Omega_X^h
     \end{aligned}.
 \end{equation*}
Therefore, 
\begin{footnotesize}
\begin{equation*}
\underbrace{\int_{x \in \Omega_X^e} \omega(x)  \dfrac{\partial  Err^{\text{w}_0}(x,\lambda_{h})}{\partial c} p(x\vert y)\mathrm{d}x}_{\textcircled{\tiny{1}}} \leq \int_{x \in \Omega_X^e} \omega^*_e  \dfrac{\partial  Err^{\text{w}_0}(x,\lambda_{h})}{\partial c} p(x\vert y)\mathrm{d}x ,
\end{equation*}
\end{footnotesize}
and 
\begin{footnotesize}
\begin{equation*}
\int_{x \in \Omega_X^h} \omega^*_h   \dfrac{\partial  Err^{\text{w}_0}(x,\lambda_{h})}{\partial c} p(x\vert y)\mathrm{d}x \geq \underbrace{\int_{x \in \Omega_X^h}  \omega(x) \dfrac{\partial  Err^{\text{w}_0}(x,\lambda_{h})}{\partial c} p(x\vert y)\mathrm{d}x }_{\textcircled{\tiny{2}}}
\end{equation*}
\end{footnotesize}
\begin{footnotesize}
\begin{equation*}
    \begin{aligned}
      \textcircled{\tiny{1}} + \textcircled{\tiny{2}} & \leq \int_{x \in \Omega_X^e} \omega^*_e  \dfrac{\partial  Err^{\text{w}_0}(x,\lambda_{h})}{\partial c} p(x\vert y)\mathrm{d}x  + \int_{x \in \Omega_X^h} \omega^*_h   \dfrac{\partial  Err'(x,\lambda_{h})}{\partial c} p(x\vert y)\mathrm{d}x \\
     \leq & \int_{x \in \Omega_X^e} \omega^*_h   \dfrac{\partial  Err^{\text{w}_0}(x,\lambda_{h})}{\partial c} p(x\vert y)\mathrm{d}x + \int_{x \in \Omega_X^h} \omega^*_h   \dfrac{\partial  Err^{\text{w}_0}(x,\lambda_{h})}{\partial c} p(x\vert y)\mathrm{d}x \\
     =&  \omega^*_h \int_{x \in \Omega_X}   \dfrac{\partial  Err^{\text{w}_0}(x,\lambda_{h})}{\partial c}\vert_{c = c^*} p(x\vert y)\mathrm{d}x \\
       = & 0
    \end{aligned}.
\end{equation*}
\end{footnotesize}
The equal relation holds if and only if
\begin{footnotesize}
\begin{equation*}
 \min \limits_{x\in \Omega^e} \omega(x)= \omega^*_e = \omega^*_h = \max \limits_{x\in \Omega^h} \omega(x).
\end{equation*}
\end{footnotesize}
Note that $\min \omega(x) < \max \omega(x)$. Therefore, $\dfrac{\partial Err^{\text{w}}(\lambda_{h}) }{\partial c} < 0$. Accordingly, the optimal complexity becomes larger. 
 \end{proof}
 
\end{appendices}



%

\end{document}